\documentclass{article}

\usepackage[eandd,final,nonatbib]{neurips_2026}
\usepackage{cite}

\usepackage[utf8]{inputenc} %
\usepackage[T1]{fontenc}    %
\usepackage{url}            %
\usepackage{microtype}      %
\usepackage{nicefrac}       %

\usepackage{graphicx}
\usepackage{booktabs}
\usepackage{multirow}
\usepackage{amsmath}
\usepackage{amssymb}
\usepackage{mathtools}
\usepackage[table]{xcolor}
\usepackage{enumitem}
\usepackage{subcaption}
\usepackage{tabularx}
\usepackage[most]{tcolorbox}
\usepackage{sidecap}
\usepackage{caption}
\tcbuselibrary{breakable,skins}

\definecolor{darkblue}{RGB}{0, 0, 130} 

\usepackage[
    colorlinks=true,
    linkcolor=magenta, 
    citecolor=darkblue,
    urlcolor=blue
]{hyperref}

\usepackage[capitalize]{cleveref}

\crefname{figure}{\textcolor{darkblue}{Fig.}}{\textcolor{darkblue}{Figs.}}
\Crefname{figure}{\textcolor{darkblue}{Fig.}}{\textcolor{darkblue}{Figs.}}

\crefname{table}{\textcolor{darkblue}{Tab.}}{\textcolor{darkblue}{Tabs.}}
\Crefname{table}{\textcolor{darkblue}{Tab.}}{\textcolor{darkblue}{Tabs.}}

\crefname{appendix}{\textcolor{darkblue}{Appx.}}{\textcolor{darkblue}{Appendix}}

\crefname{section}{\textcolor{darkblue}{Sec.}}{\textcolor{darkblue}{Secs.}}
\Crefname{section}{\textcolor{darkblue}{Section}}{\textcolor{darkblue}{Sections}}

\crefname{appsection}{\textcolor{darkblue}{Appx. Sec.}}{\textcolor{darkblue}{Appx. Secs.}}
\Crefname{appsection}{\textcolor{darkblue}{Appx. Section}}{\textcolor{darkblue}{Appx. Sections}}

\crefname{appfigure}{\textcolor{darkblue}{App.~Fig.}}{\textcolor{darkblue}{App.~Figs.}}
\Crefname{appfigure}{\textcolor{darkblue}{App.~Fig.}}{\textcolor{darkblue}{App.~Figs.}}

\crefname{apptable}{\textcolor{darkblue}{App.~Tab.}}{\textcolor{darkblue}{App.~Tabs.}}
\Crefname{apptable}{\textcolor{darkblue}{App.~Tab.}}{\textcolor{darkblue}{App.~Tabs.}}

\usepackage{xspace}
\newcommand{\mtgad}{MTG-AD\xspace}
\newcommand{\mtgadeval}{MTG-AD-Eval\xspace}
\newcommand{\mtgcodeurl}{\href{https://github.com/google/mtg-ad}{github.com/google/mtg-ad}}
\newcommand{\mtgdataurl}{\href{https://hf.co/datasets/google/mind-the-gap-inline-ad}{hf.co/datasets/google/mind-the-gap-inline-ad}}

\definecolor{firstplace}{HTML}{E2E2E2}
\definecolor{secondplace}{HTML}{F2F2F2}

\title{Mind The Gap (MTG-AD): Dataset and Fine-grained Evaluation for End-to-End Inline Audio Descriptions}

\author{%
  Subhashini Venugopalan\textsuperscript{1}, 
  Jenny Tan\textsuperscript{1}, 
  Taylor Roper\textsuperscript{1}, 
  Jimmy Tobin\textsuperscript{1}, \\
  \textbf{Anton Kast\textsuperscript{1}}, 
  \textbf{Alicia Martin\textsuperscript{1}}, 
  \textbf{Sam Sepah\textsuperscript{1}}, 
  \textbf{Amy Pavel\textsuperscript{1,2}} \\
  \textsuperscript{1}Google Research, \textsuperscript{2}UC Berkeley %
}

\begin{document}

\maketitle

\begin{abstract}
Audio descriptions (AD) are essential for making visual media accessible to blind and low-vision (BLV) users. While Multimodal Large Language Models (MLLMs) offer a scalable solution for generating audio descriptions on-demand, their performance on ``inline'' audio descriptions -- which must fit within existing silences in a video -- remains under-explored, particularly for diverse user-generated content (UGC). We present a comprehensive investigation into MLLM-generated inline audio descriptions. First, we introduce \textbf{\mtgad}, a professionally annotated dataset of 36 hours of videos, each averaging $\sim$5 mins. and 8k+ descriptions in total, co-created by paired blind and sighted AD professionals to provide a high-quality benchmark for this task. Second, we propose \textbf{\mtgadeval}, an evaluation framework based on audio description expert guidelines that provides fine-grained, reference-free actionable metrics for model improvement. Crucially, our framework evaluates the end-to-end task: it renders the generated scripts into audio tracks to assess their fit within the original video's natural silences, minimizing disruptive overlaps. Our analysis across frontier MLLMs reveals that while models accurately describe visual events and substantially reduce speech overlaps, gaps persist in narrative pacing and flow.
Finally, we empirically validate that our automated and reference-free metrics correlate significantly with human quality ratings from sighted raters and 7 BLV participants, finding that while users narrowly prefer human-authored audio descriptions, they find MLLM-generated audio descriptions highly beneficial for comprehension.
We provide actionable system design guidance and release \mtgad at \mtgdataurl \ and \mtgadeval at \mtgcodeurl. %
\end{abstract}

\section{Introduction}
\label{sec:intro}

\begin{figure*}[!tbh]
    \vspace{-0.4cm}
    \centering
    \includegraphics[width=\linewidth]{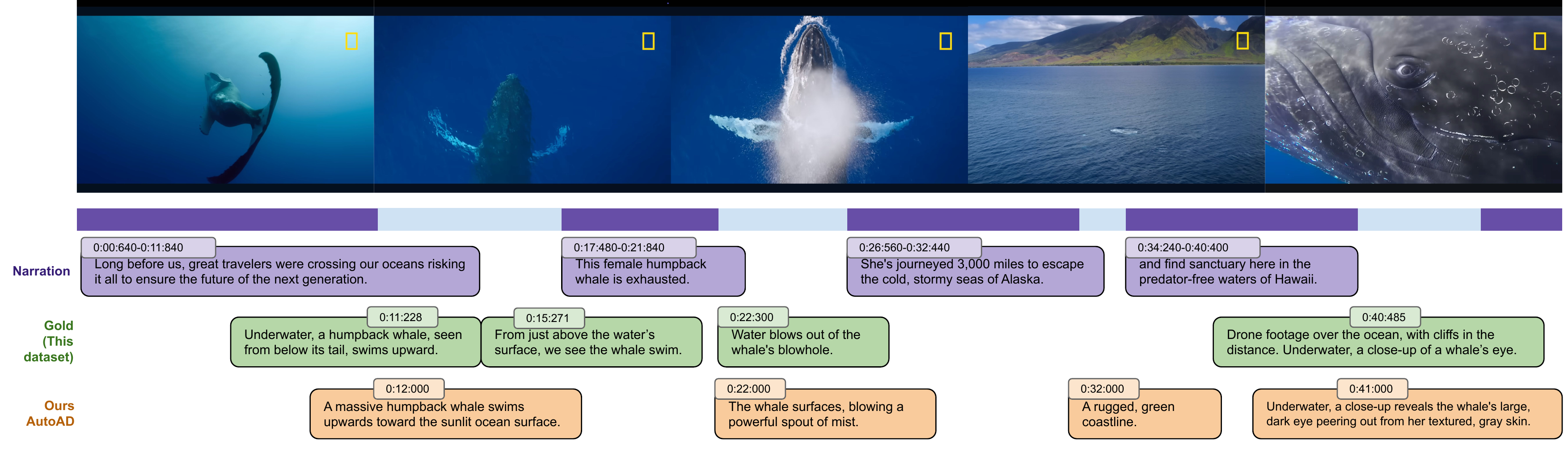}
    \caption{
    \small\textbf{Temporal Constraints of Inline Audio Descriptions.} Inline AD must fit precisely into natural silences within a video's narration (purple). 
     Both the professional ``Gold'' human AD from \mtgad (green) and our model-generated ``AutoAD'' (orange) fit within these tight temporal gaps without overlapping speech. While AutoAD generates vivid, coherent descriptions with timing-aware prompts, the professional AD achieves stronger narrative flow by capturing camera perspectives (\textit{e.g.}, ``drone footage'').
     }
    \label{fig:dataset_examples}
    \vspace{-0.5cm}
\end{figure*}

Blind and low-vision (BLV) viewers turn to user-generated videos to learn new skills, pursue hobbies, and stay informed~\cite{liu2021what}. Yet creators rarely verbalize essential visual details, leaving many videos hard to follow~\cite{liu2021what,liu2022crossa11y}. Audio descriptions (AD) make videos accessible by narrating key visual events to complement the existing audio~\cite{blind2020american,w3c2022audio}. Unlike extended AD that pauses playback, \emph{inline} AD preserves the video's runtime by fitting descriptions into natural pauses (\cref{fig:dataset_examples}). Effective inline AD thus goes beyond listing what is on screen: it requires \textit{temporal precision} and \textit{content selectivity} to build a \textit{coherent narrative} that enriches, rather than repeats or interrupts, the original audio.

While high-quality audio descriptions are professionally crafted for film and TV~\cite{packer2015overview}, it is rarely available for user-generated videos~\cite{liu2021what}, motivating semi-automated~\cite{liu2022crossa11y,pavel2020rescribe,yuksel2020human,gagnon2009towards} and automated generation approaches~\cite{wang2021toward,lin2023mm,chu2024llm,xie2024autoad,han2023autoad,li2025videoa11y} that have the potential to scale access for millions of BLV audience members.
Recent progress in general-purpose multimodal large language models (MLLMs) makes it newly possible to generate AD scripts directly from video. Models such as Gemini 2.5/3.5~\cite{google2025gemini25}, GPT-5.2~\cite{openai2025gpt52}, and Qwen-3.5-Omni~\cite{qwen2026qwen35omni} can take a video as input and output a timestamped narration script to be rendered via text-to-speech (TTS). Yet \textbf{\textit{are MLLM-generated audio descriptions understandable and enjoyable for BLV audiences of user-generated videos?}} Answering this requires evaluating the full end-to-end pipeline: generating the AD script, synthesizing speech via TTS, and measuring how well the rendered audio fits within the natural silences of the original video.

Prior work largely evaluates generated AD on trimmed video segments against a single reference~\cite{han2024autoad,yue2023movie101}, through short-clip human ratings, or with coarse LLM scores~\cite{li2025videoa11y}. Evaluating isolated clips out of context misses sequence-level coherence~\cite{khandelwal2025more}, spoken AD duration, un-narrated ``dead air'' gaps, and speech overlaps that shape the end-to-end BLV listening experience. Moreover, unlike movies with professional reference AD~\cite{rohrbach2017movie}, user-generated videos rarely have expert annotations, forcing prior benchmarks to rely on synthetic~\cite{han2024autoad} or novice-authored proxies where coarse metrics quickly saturate~\cite{li2025videoa11y}.

Our work makes three contributions: First, we \textit{establish a high-quality benchmark for inline AD on user-generated videos}, \mtgad, comprising professionally crafted ADs for 433 YouTube videos (${\sim}36$ hours) co-created by paired blind and sighted AD specialists, alongside outputs from nine frontier MLLMs. Second, we introduce \mtgadeval, an \textit{end-to-end, reference-free evaluation framework} grounded in expert AD guidelines that pairs TTS-based temporal metrics with multimodal LLM judges, analyzing cross-judge biases and transfer to YouDescribe. Finally, we \textit{validate our metrics and benchmark against human judgments} from sighted raters and 7 BLV participants, showing strong alignment with human ratings, including resolving reference-matching constraints in prior works, and identifying where frontier MLLMs still trail human narrative pacing.

\vspace{-0.2cm}
\section{Related Work}
\vspace{-0.2cm}
\begin{table*}[b]
\vspace{-0.3cm}
\centering
\resizebox{\textwidth}{!}{%
\begin{tabular}{@{}llcccccc@{}}
\toprule
\textbf{Dataset} & \textbf{Video Domain} & \textbf{Total Duration} & \textbf{Avg.\ Clip / Video Len.} & \textbf{Source of AD} & \textbf{\# AD Segments} & \textbf{Speech / Audio} & \textbf{Eval.\ Scope} \\
\midrule
\textbf{M-VAD}~\cite{rohrbach2017movie} & Movies & 84.6 hrs (92 movies) & 6.2\,s & Professionals (DVS) & 48,986 & Mixed & Segment (gaps) \\
\textbf{MAD}~\cite{soldan2022mad} & Movies (CLIP feats.) & 1,207.3 hrs (650 movies) & 4.1\,s & Professionals (AudioVault) & 384,600 & Mixed & Segment (gaps) \\
\textbf{MAD-v2}~\cite{han2023autoad} & Movies (CLIP feats.) & 892 hrs (488 movies) & ${\sim}4.1$\,s & Professionals (AudioVault) & 264,000 & Mixed & Segment (gaps) \\
\textbf{Movie101v2}~\cite{yue2023movie101} & Movies (ZH/EN) & 353 hrs (203 movies) & 11--20\,s & Professionals & 71,000 & Non-dialogue & Segment (gaps) \\
\textbf{CMD-AD}~\cite{han2024autoad} & Movie Clips & 477 hrs (1,432 movies) & ${\sim}2$\,min & Professionals (AudioVault) & ${\sim}101,000$ & Mixed & Segment (gaps) \\
\textbf{TV-AD}~\cite{xie2024autoad} & TV Series & 251 hrs (426 episodes) & 2.1\,s & Professionals (AudioVault) & 34,013 & Mixed & Segment (gaps) \\
\midrule
\textbf{HowTo-AD}~\cite{han2024autoad} & Instructional & 23,652 hrs (${\sim}180$k videos) & Variable & Synthetic (ASR-adapted) & 3.4M & Heavy dialogue & N/A (train only) \\
\textbf{VideoA11y-40K}~\cite{li2025videoa11y} & User-generated & -- (40k videos) & ${\sim}10$\,s & MLLMs (Synthetic) & 40,000 & Ignored & Segment \\
\textbf{YouDescribe (YuWA)}~\cite{institute2022youdescribe} & YouTube & ${\sim}310$ hrs (${\sim}5.5$k videos) & ${\sim}5$\,min & Volunteers / Novices & ${\sim}76,000$ & Mixed & N/A (archive) \\
\midrule
\textbf{\mtgad (Ours)} & \textbf{User-generated} & \textbf{36 hrs (433 videos)} & \textbf{${\sim}5$\,min} & \textbf{Professional + Blind pair} & \textbf{8,126} & \textbf{Mixed} & \textbf{Full (end-to-end)} \\
\bottomrule
\end{tabular}%
}
\caption{\textbf{Comparison of Audio Description Datasets.} Prior movie and TV benchmarks evaluate isolated silent intervals using pre-trimmed clips or pre-extracted visual features, while prior user-generated or instructional datasets rely on volunteer narrations, short clips with ignored audio, or synthetic ASR adaptations. \mtgad provides full-length user-generated videos co-annotated by paired sighted and blind AD professionals and evaluates end-to-end inline AD across each entire video.}
\label{tab:dataset_comparison}
\vspace{-0.3cm}
\end{table*}
\textbf{Audio Description Datasets.}
Most existing AD benchmarks draw from professional descriptions created for movies~\cite{rohrbach2017movie,soldan2022mad,han2023autoad,yue2023movie101,han2024autoad} or TV series~\cite{xie2024autoad} (\cref{tab:dataset_comparison}). While these corpora establish what and how to describe in scripted entertainment, user-generated videos vary widely in intent, camera work, editing pace, on-screen text, and speech density, from silent montages to near-continuous narration. Movie AD benchmarks also evaluate models on pre-trimmed 2--20\,s clips around known AD timestamps, whereas inline AD for user-generated content requires reasoning over untrimmed, multi-minute videos (averaging 5 minutes in \mtgad, a horizon shared with long-video benchmarks such as Neptune~\cite{nagrani2024neptune}) to decide both \emph{when} and \emph{what} to describe. Because professional AD rarely exists for user-generated videos, prior datasets rely on volunteer descriptions of uneven quality on YouDescribe~\cite{institute2022youdescribe}, synthetic MLLM descriptions of short ${\sim}10$\,s clips~\cite{li2025videoa11y}, or repurposed narrations from instructional videos~\cite{han2024autoad}. \mtgad addresses this gap with 36 hours of full-length user-generated videos co-annotated by paired sighted and blind AD professionals.

\begin{figure*}[!t]
\vspace{-0.4cm}
    \includegraphics[width=\textwidth]{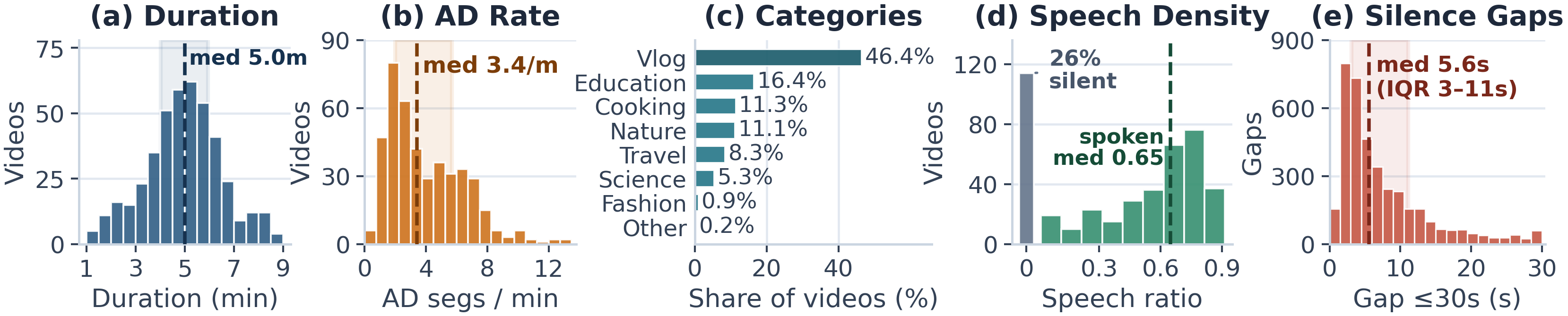}
    \caption{\textbf{Dataset Statistics.} \mtgad comprises 433 professionally annotated videos (${\sim}36$ hours): (a) video durations center around 5 minutes; (b) human-annotated AD segments per minute; (c) vlogs form the largest category alongside travel, nature, and education; both (d) speech density and (e) silence-gap distribution illustrate the tight pauses into which inline AD must fit.}
    \label{fig:dataset_stats}
    \vspace{-0.6cm}
\end{figure*}

\textbf{Evaluating Audio Descriptions.}
Automated AD evaluation largely compares generated text against a ground-truth reference~\cite{gao2025audio}, using n-gram or embedding similarity metrics such as BLEU~\cite{papineni2002bleu} and BERTScore~\cite{zhang2019bertscore}, or domain-specific metrics for character naming and semantic alignment such as MNScore~\cite{yue2023movie101}, CRITIC~\cite{han2024autoad}, and LLM-AD-Eval~\cite{han2024autoad}. However, single-reference similarity on isolated clips poorly reflects listener comprehension and penalizes valid alternative descriptions. Recent work moves beyond isolated single-reference scoring toward sequence-level coherence~\cite{khandelwal2025more} and QA-based comprehension checks~\cite{kala2024what}, yet still evaluates text scripts without rendering audio to test whether descriptions fit natural pauses without overlapping speech or leaving long silences.

Human studies with external raters or BLV listeners better capture end-to-end playback quality, though full-video listening studies are time-consuming and typically cover few videos~\cite{branje2012livedescribe,pavel2020rescribe}. On short ${\sim}10$\,s clips, VideoA11y~\cite{li2025videoa11y} found that guideline-prompted MLLM descriptions surpassed novice captions and matched trained describers, and that LLM-as-a-judge ratings aligned with BLV judgments, echoing broader findings on LLM judges for audio captioning~\cite{wu2024claira}. Still, prior work has not evaluated end-to-end inline AD on full-length user-generated videos, compared MLLM outputs against professionally co-authored AD in that setting, or validated reference-free timing and quality metrics against sighted and BLV listeners.

\vspace{-0.25cm}
\section{Mind the Gap (\mtgad) Dataset}
\label{sec:dataset}
\vspace{-0.25cm}
\mtgad benchmarks end-to-end inline audio description on user-generated YouTube videos, a domain frequently watched by BLV viewers~\cite{liu2021what,institute2022youdescribe,li2025videoa11y} where professional AD rarely exists (\cref{fig:dataset_examples,fig:dataset_stats}). The dataset comprises 433 English videos (${\sim}35.9$ hours; mean $5.0$\,min) partitioned into 332 validation videos for prompt and method development and 101 test videos for multi-model evaluation. Each video includes inline AD co-authored by paired sighted and blind AD professionals, a timestamped YouTube ASR transcript ($26.3\%$ of videos have no transcribed speech, such as music-backed montages), and pre-extracted dense visual event lists ($\mathcal{D}$, \cref{sec:metrics}). For the 101 test videos, we also release unedited AD scripts and per-video metric scores across 12 model--prompt configurations spanning four prompting strategies and nine frontier multimodal models.

\vspace{-0.2cm}
\subsection{Video Selection and Filtering}
\vspace{-0.2cm}
We sampled 557 candidate English videos around 3--7 minutes long from YouTube channels with at least 1,000 subscribers across diverse user-generated categories, led by vlogs (46\%), education (16\%), cooking, nature, travel, and science in the final set (\cref{fig:dataset_stats}c). The professional AD teams reviewed every candidate in full to assess whether inline AD could be added without pausing playback, filtering candidates on two criteria:
(i)~\textbf{Temporal viability}: videos with nonstop dialogue or narration, or vlogs whose on-screen subtitles left no audio space for additional descriptions, were marked non-viable, though annotators revisited borderline cases to include partial descriptions where brief pauses permitted; and
(ii)~\textbf{Topical diversity}: after annotating 3--5 videos of a recurring style (\textit{e.g.}, overhead cooking tutorials), further near-duplicates were skipped.
Of the 557 candidates, 438 ($78.6\%$) were annotated, 92 ($16.5\%$) were excluded as non-viable, and 27 ($4.8\%$) were skipped for redundancy. Five duplicate candidates were annotated twice; retaining the first annotation yields 433 unique videos totaling $35.9$ hours (\cref{fig:dataset_stats}a,d).

\vspace{-0.2cm}
\subsection{Professionally Co-Authored AD}
\vspace{-0.1cm}
Descriptions were produced by two two-person professional teams, each pairing one sighted and one blind AD writer so that every script reflects both visual detail and BLV listener priorities. For each video, the sighted writer first watched the full video to establish narrative context and pacing, then rewatched it in segments to draft concise, timestamped descriptions of key visual events and on-screen text not already conveyed by speech or sound effects. Next, the writer imported the script into an audio-editing environment against the original soundtrack, adjusting phrasing and start times so each spoken description fit cleanly into natural pauses without overlapping dialogue. Finally, the blind writing partner reviewed the timed playback end-to-end to audit narrative clarity and standalone comprehension before sign-off. Across all 433 videos, \mtgad contains 8,126 AD segments (6,214 validation, 1,912 test), with a mean of 18.8 (median 16, IQR 10--26) segments per video, 17.0 words per segment (median 15), and a median rate of 3.4 segments per minute (IQR 2.0--5.7; \cref{fig:dataset_stats}b).

\vspace{-0.2cm}
\subsection{Eliciting Model-Generated AD on Untrimmed Videos}
\vspace{-0.2cm}

\begin{figure*}[t]
\vspace{-0.2cm}
    \centering
    \includegraphics[width=\linewidth]{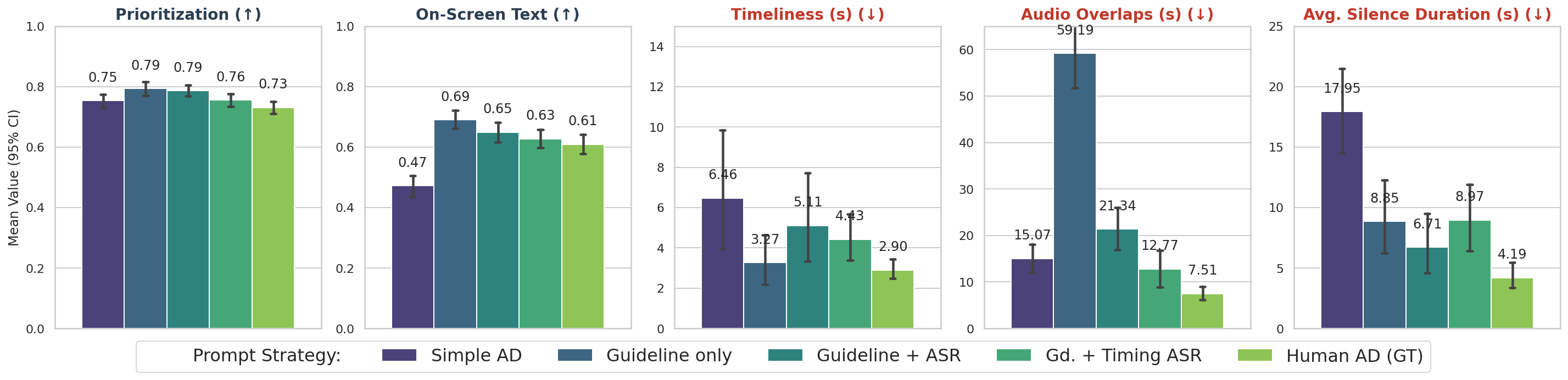}
    \caption{\textbf{Prompt strategy comparison} on content and temporal metrics (\cref{sec:metrics}) for Gemini-2.5 Flash. \textbf{Content} (first two panels; higher is better): guideline prompts improve visual prioritization and on-screen text recall over \textit{Simple AD}. \textbf{Temporal precision} (last three panels; lower is better): \textit{Simple AD} leaves long silences, whereas \textit{Guideline only} overfills gaps and overlaps speech; \textit{Gd.\ + Timing ASR}, which adds the timestamped transcript and per-gap word budgets, best balances content against timing.}
    \label{fig:prompt_comparison_main}
    \vspace{-0.4cm}
\end{figure*}

Unlike trimmed-clip captioning with pre-segmented gaps, generating inline AD on untrimmed videos requires a model to select salient visual events while aligning to speech pauses and fitting spoken AD within each gap.
To study how well MLLMs handle this end-to-end task and what context or reasoning steps resolve their failure modes, we designed four prompts that we studied on Gemini-2.5 Flash.
\textbf{Simple AD} (\cref{fig:prompt_basic}) tests minimal-instruction capability, requesting only timestamped descriptions during non-speech intervals;
\textbf{Guideline only} (\cref{fig:prompt_guideline_ad}) adds expert AD rules (\textit{e.g.}, prioritizing key visuals, reading on-screen text verbatim, remaining objective, and omitting audible cues), testing whether domain guidelines suffice when speech pauses must be inferred directly from video;
\textbf{Guideline + ASR} (\cref{fig:prompt_asr_guideline}) supplies the timestamped speech transcript (start and end times) and a $1$--$3$ words/s speaking-rate target to ground pause boundaries in text; and
\textbf{Gd.\ + Timing ASR} (\cref{fig:prompt_timing_asr}) decomposes temporal planning into explicit steps: computing each silent gap's duration, capping length at $3$ words per silent second (at most three sentences), and skipping details when space is insufficient.
To compare model families under our strongest elicitation setup (\textbf{Gd.\ + Timing ASR}), on the 101 test videos we pair all four Gemini-2.5 Flash prompt runs with eight additional frontier models: GPT-4.1 mini, GPT-5.2, Claude 4, Claude 4.5, Qwen-2-VL-Max, Gemini-3.5 Flash, and Qwen-3.5-Omni-Flash/Plus.  Providing the timestamped transcript alongside the video gives both native audio-video models (Gemini-2.5/3.5 Flash, Qwen-3.5-Omni) and vision-and-text models identical speech content and silence boundaries without requiring raw audio ingestion.

\vspace{-0.2cm}
\section{\mtgadeval: Evaluation Framework}
\label{sec:metrics}
\vspace{-0.2cm}
\textbf{From expert guidelines to metrics.} Following prior work~\cite{pavel2020rescribe,li2025videoa11y}, we reviewed five expert AD guidelines chosen for breadth and comprehensiveness: the American Council of the Blind~\cite{blind2020american}, DCMP's Description Key~\cite{dcmp2020description}, ADLab's film guide~\cite{remael_pictures}, W3C's WCAG~\cite{w3c_g8}, and YouDescribe's tips~\cite{institute2022youdescribe}. We also reviewed studies of BLV audiences' perceptions of AD to find where audience preferences depart from prescribed rules~\cite{natalie2024audio,liu2021what,jiang2023beyond,jiang2024s}. Together with researchers in HCI, accessibility, and computer vision, we distilled these to ten guidelines covering content (G1--G4), style (G5--G6), and audio (G7--G10); \cref{appx_sec:full_eval_metrics} lists them all. In a pilot study, BLV audience members reviewed AD generated with a basic prompt and named the problems that most hurt their experience: AD talking over speech, long unexplained silences, and descriptions out-of-sync. We therefore turn the six guidelines tied to these problems (G1.1--G1.3, G2, G7, G8) into automatic metrics, and add the VideoA11y rubrics~\cite{li2025videoa11y} to rate writing quality. \Cref{tab:metric_summary} summarizes the metrics.

\textbf{End-to-end setup and notation.} We score the AD of the whole video as a listener would hear it, not isolated clips. For a video of duration $T$ (seconds), a system outputs an AD script $\{(s_i, x_i)\}_{i=1}^{N}$ with segment start times $s_i$ and texts $x_i$. We synthesize each $x_i$ with an off-the-shelf text-to-speech (TTS) engine at a fixed $1.2\times$ speaking rate to obtain its spoken duration $d_i$ and playback interval $I_i=[s_i,\,s_i+d_i]$. Let $A=\bigcup_{i=1}^{N} I_i$ denote the union of AD intervals and $|\cdot|$ the total duration of a set of intervals, so $|A|$ is the total time AD plays. The metrics use two precomputed inputs per video that require no human-written AD:
(i)~the timestamped ASR transcript, yielding both spoken words and speech intervals $[a_j,b_j]$ with union $S=\bigcup_j [a_j,b_j]$; and
(ii)~\emph{dense visual information} $\mathcal{D}$, a timestamped list of the video's important visual events and on-screen text, generated by Gemini-2.5 Flash (prompt in \cref{fig:prompt_vis_info}).
Combined with \texttt{Quality-RF}, which judges scripts directly against the video and transcript, every metric except \texttt{VideoA11y-Ref} runs on unseen videos without human references. We used Gemini-2.5 Flash as the default judge (and switch to Gemini 3.5 Flash in the released code) and analyze cross-judge sensitivity in \cref{sec:experiments,sec:llm_judge}.

\begin{table}[!t]
\centering
\caption{\textbf{Evaluation metrics at a glance.} Metrics are computed per video and averaged across videos. \emph{Scorer}: Prog.\ = programmatic or word arithmetic; LLM = text-only LLM judge; MLLM = video + text judge (prompts in \cref{appx_sec:prompts_ad_eval}). \emph{Ref-free}: runs without human AD. $\mathcal{D}$ = dense visual information; TTS = AD rendered at $1.2\times$.}
\label{tab:metric_summary}
\footnotesize
\setlength{\tabcolsep}{3.5pt}
\renewcommand{\arraystretch}{1.12}
\begin{tabularx}{\textwidth}{@{}>{\raggedright\arraybackslash}p{2.6cm}>{\raggedright\arraybackslash}Xccc@{}}
\toprule
\textbf{Metric (guideline)} & \textbf{Question it answers $\rightarrow$ computation} & \textbf{Scorer} & \textbf{Ref-free} & \textbf{Better} \\
\midrule
\multicolumn{5}{@{}l}{\cellcolor{secondplace}\textit{Content metrics}} \\
Prioritization (G1.1) & Is AD spent on salient visual events? $\rightarrow$ Judge scores each segment as $c_i$: covered (1), partly (0.5), or not (0) for support in $\mathcal{D}$; mean $\bar c$. & LLM & \checkmark & $\uparrow$ $[0,1]$ \\
No Redundancy (G1.2) & Does AD avoid restating audible cues? $\rightarrow$ Judge scores each segment's overlap with nearby transcript or explicit sound cues, $r_i\in[0,1]$; $1-\bar r$. & LLM & \checkmark & $\uparrow$ $[0,1]$ \\
On-Screen Text (G1.3) & Is on-screen text conveyed? $\rightarrow$ Fraction of on-screen text words from $\mathcal{D}$ in the combined AD and ASR vocabulary. & Prog. & \checkmark & $\uparrow$ $[0,1]$ \\
\midrule
\multicolumn{5}{@{}l}{\cellcolor{secondplace}\textit{Temporal metrics}} \\
Timeliness (G2) & Is each description heard near its visual event? $\rightarrow$ Judge matches each AD segment to an event at $t_i$ in $\mathcal{D}$; mean $|s_i - t_i|$ on matches. & LLM & \checkmark & $\downarrow$ sec. \\
Audio Overlaps (G7) & Does AD talk over speech? $\rightarrow$ Total duration $|A \cap S|$ where TTS-rendered AD and speech play simultaneously. & Prog. & \checkmark & $\downarrow$ sec. \\
Audio Coverage (G8) & How much available silence does AD use? $\rightarrow$ $|A| / (T - |S|)$; values ${>}1$ indicate AD exceeds total non-speech time. & Prog. & \checkmark & $\uparrow$ ($\le 1$) \\
Avg.\ Silence (G8) & How long are the remaining quiet stretches? $\rightarrow$ Mean duration of gaps in $[0,T]\setminus(A\cup S)$ with neither speech nor AD. & Prog. & \checkmark & $\downarrow$ sec. \\
\midrule
\multicolumn{5}{@{}l}{\cellcolor{secondplace}\textit{Holistic script quality (VideoA11y rubric: Descriptive, Objective, Accurate, Clear; 4 scores)}} \\
\texttt{VideoA11y-Ref} \newline \cite{li2025videoa11y} & How closely does AD match human AD? $\rightarrow$ Judge rates the script on each dimension against the human-written reference. & LLM & $\times$ & $\uparrow$ 1--5 \\
\texttt{Quality-RF} (ours) & How well does AD describe the video? $\rightarrow$ Same four ratings, judged directly against the video and its ASR transcript. & MLLM & \checkmark & $\uparrow$ 1--5 \\
\bottomrule
\end{tabularx}
\vspace{-0.3cm}
\end{table}

\vspace{-0.2cm}
\subsection{Content Metrics: What the AD Says}
\label{sec:content_metrics}
\vspace{-0.1cm}
\textbf{Prioritization (G1.1)} asks whether the AD spends its limited budget on visual information needed to follow the video. Given the AD script and $\mathcal{D}$, a judge scores each segment with $c_i\in\{0, 0.5, 1\}$ for whether $x_i$ describes salient content in $\mathcal{D}$ (\cref{fig:prompt_coverage}). We report $\frac{1}{N}\sum_i c_i$, which measures segment-level precision against $\mathcal{D}$. It penalizes trivial or invented details but not omissions (measured partly in onscreen-text) or under-description (captured by Coverage and Avg.\ Silence).
\textbf{No Redundancy (G1.2)} checks that AD does not restate what a listener can already hear, such as dialogue or an obvious sound (e.g., door slam). We interleave the transcript and the AD by time, and a judge scores each segment's redundancy $r_i\in\{0,0.5,1\}$ (\cref{fig:prompt_redundancy}). We report $1-\frac{1}{N}\sum_i r_i$, so higher is better.
\textbf{On-Screen Text (G1.3)} measures whether text shown in the video (signs, titles, captions) is conveyed to the listener. Let $W$ be the words of the on-screen text listed in $\mathcal{D}$ and $V$ the set of words spoken in the AD or the transcript. The score is the fraction of $W$ found in $V$, which is a word-level recall computed programmatically (an initial LLM-judged version is shown in ~\cref{fig:prompt_onscreen}). Videos without on-screen text are excluded.

\vspace{-0.2cm}
\subsection{Temporal Metrics: When and How Much the AD Speaks}
\vspace{-0.1cm}
\textbf{Timeliness (G2)} measures how far each description is from the moment its event appears. A judge pairs each AD segment's start time $s_i$ with the event in $\mathcal{D}$ it describes, occurring at time $t_i$, or marks it unmatched (\cref{fig:prompt_timeliness}). Over the set $M$ of matched segments, we report Timeliness  %
in seconds  (unmatched segments are penalized by G1.1).
While Timeliness evaluates when AD starts, fitting inline AD into pauses depends on the rendered speech interval $I_i$ so it shouldn't spill into the next dialogue.
\textbf{Audio Overlaps (G7)} is the total time during which AD and speech play together, $T_{\text{overlap}} = |A \cap S|$.
\textbf{Audio Coverage (G8)} compares how long the AD speaks relative to how much silence is available $\text{Coverage} = {|A|} / {T - |S|}$. 
This checks if the AD covers most of the video without filling every pause, so values somewhat below $1$ are desired. We deliberately do not clip the ratio. A value above $1$ means rendered AD exceeds all available silence combined (\textit{e.g.}, $1.23$ for the \textit{Guideline only} prompt), so it flags over-generation.
These three metrics are computed as:
\begin{equation*}
\text{Timeliness} = \tfrac{1}{|M|}\textstyle\sum_{i\in M} |s_i - t_i|; %
\quad
T_{\text{overlap}} = \big|\bigcup_i I_i \,\cap\, \bigcup_j [a_j,b_j]\big|; 
\quad
\text{Coverage} = \frac{\big|\bigcup_i I_i\big|}{T - \big|\bigcup_j [a_j,b_j]\big|}.
\end{equation*}
\textbf{Avg.\ Silence (G8)} measures the dead air left after the AD is inserted. We take the gaps in $[0,T]$ during which neither speech nor AD plays, including those at the start and end, and report their mean length in seconds. Coverage and Avg.\ Silence complement each other: a script can reach high Coverage with a few long segments yet leave long silences elsewhere.

\vspace{-0.2cm}
\subsection{Holistic Script Quality}
\vspace{-0.1cm}
Following VideoA11y~\cite{li2025videoa11y}, an LLM judge rates the full AD script on a likert scale (1--5) on four dimensions: \textbf{Descriptive} (vivid, relevant visual detail), \textbf{Objective} (reports what is visible without speculation), \textbf{Accurate} (visual details are correct), and \textbf{Clear} (easy to follow without seeing the video). We evaluate two variants:
\begin{itemize}[nosep, leftmargin=*]
    \item \textbf{\texttt{VideoA11y-Ref}} (\cref{fig:prompt_videoa11y}) is the original reference-based setup: the judge compares the generated AD with the human-written AD. A 5-minute video admits many valid descriptions, so its \emph{Accurate} score effectively measures \emph{Reference Alignment}, which penalizes correct details that the human describer happened not to mention. Human AD scored this way is compared with itself and therefore serves only as an upper bound.
    \item \textbf{\texttt{Quality-RF}} (ours, \cref{fig:prompt_videoa11y_rf}) removes the reference human AD. The judge is given both the raw video (frames and audio) and the ASR transcript to ground spoken dialogue explicitly (input ablations in \cref{appx_sec:metric_validation}); it summarizes the video, explains its reasoning, and then scores the same four dimensions. This avoids the reference-matching penalty and works without human AD.
\end{itemize}

\vspace{-0.2cm}
\section{Experiments and Results}
\label{sec:experiments}
\vspace{-0.2cm}
We evaluate (1) prompting strategies and frontier MLLMs with our automated metrics, (2) the reliability of LLM judges and transfer to the YouDescribe dataset~\cite{institute2022youdescribe}, and (3) human ratings from sighted raters and BLV participants alongside agreement with our metrics. Each value is a mean over videos with a 95\% confidence interval. Full tables are shown in \cref{appx_sec:full_results}.
\vspace{-0.2cm}
\subsection{Metric-based Evaluations}
\vspace{-0.1cm}

\begin{figure*}[!t]
\vspace{-0.2cm}
    \centering
    \includegraphics[width=\linewidth]{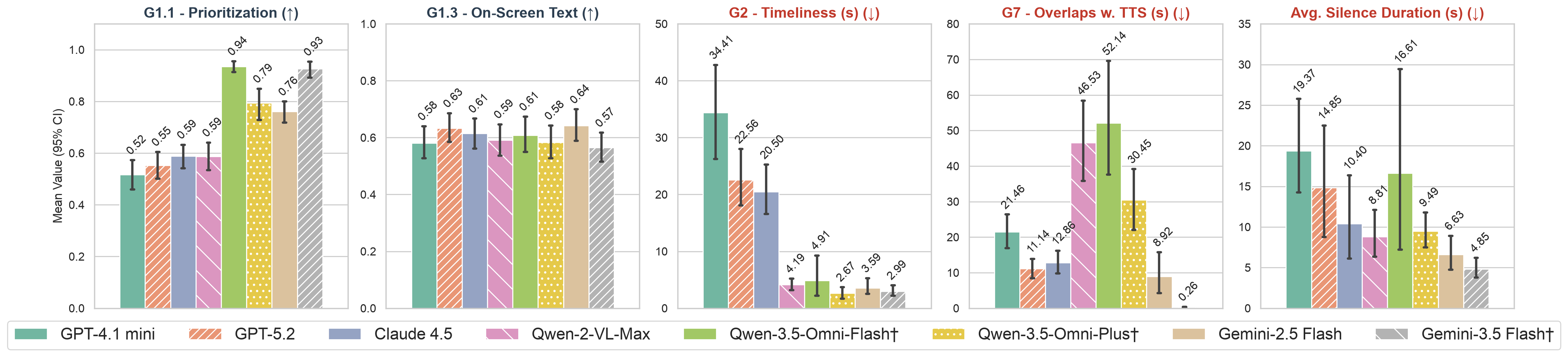}
    \caption{\textbf{Frontier model comparison} on content (G1.1, G1.3; higher is better) and temporal metrics (G2, G7, Avg.\ Silence; lower is better), all with the Gd.\ + Timing ASR prompt. Bars show means with 95\% CIs over the 101 test videos. Models marked $^{\dagger}$: G1.1 and G2 scored by a Gemini-3.5 Flash judge (others: Gemini-2.5 Flash). \cref{tab:main_res_model_comp,tab:omninew_models_comp} \ show detailed scores. %
    }
    \label{fig:model_comparison_main}
    \vspace{-0.3cm}
\end{figure*}

\begin{figure*}[!t]
    \centering
    \includegraphics[width=\linewidth]{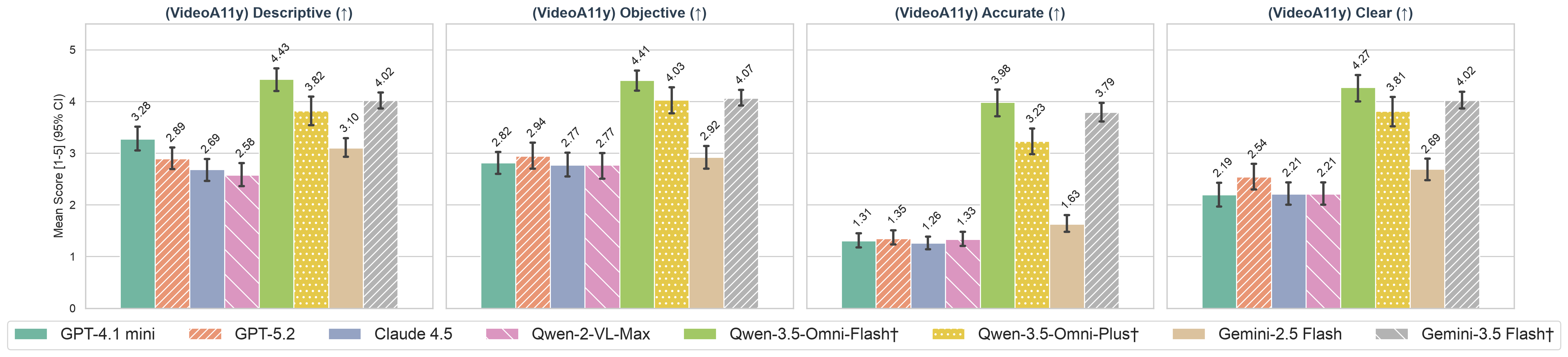}
    \caption{\textbf{Qualitative alignment (\texttt{VideoA11y-Ref}) ratings} on the same videos as \cref{fig:model_comparison_main}. Because the judge compares against the human AD, \emph{Accurate} here measures agreement with that reference (Reference Alignment). Models marked $^{\dagger}$ were scored by a Gemini-3.5 Flash judge and the others by Gemini-2.5 Flash, so scores should be compared only within each group.}
    \label{fig:model_comparison_videoally}
\vspace{-0.7cm}
\end{figure*}

\textbf{Metric-guided prompt refinement.} Scoring the four Gemini-2.5 Flash prompts across validation videos (\cref{fig:prompt_comparison_main}, \cref{tab:main_results}) illustrates how \mtgadeval diagnoses failure modes to guide prompt design. \textit{Simple AD} under-describes the video (Coverage $0.41$, Avg.\ Silence $17.95$\,s, On-Screen Text $0.47$). Adding domain rules in \textit{Guideline only} raises Prioritization ($0.79$) and On-Screen Text ($0.69$), yet \mtgadeval exposes severe over-generation ($1.23$ Coverage, $59.19$\,s of speech overlap) that content metrics alone would miss. With \textit{Guideline + ASR} which includes explicit time spans of dialogue overlaps reduce to $21.34$\,s. Further emphasizing per-gap duration and word-budget constraints in \textit{Gd.\ + Timing ASR} reduces overlaps considerably to $12.77$\,s while preserving Prioritization ($0.76$). Human AD remains better timed ($7.51$\,s overlap, $4.19$\,s Avg.\ Silence); its slightly lower Prioritization ($0.73$) arises because $\mathcal{D}$ itself is MLLM-extracted, leaving fine-grained human-observed actions (\textit{e.g.}, a truck braking or swerving in \cref{fig:dataset_examples_appendix}) and brief pause-fitted beats (e.g. A seal blinking) uncredited.

\textbf{Frontier models.} We score nine models on the 101 test videos (\cref{fig:model_comparison_main}, \cref{tab:main_res_model_comp,tab:omninew_models_comp}). 
Native audio-video models (and more recent versions) lead on Prioritization: Qwen-3.5 and Gemini models have scores between $0.76$--$0.94$, versus $0.51$--$0.59$ for the others.
Timeliness follows a similar split: with the recent Qwen-3.5 and Gemini-3.5 models describing events within $2-5$\,s of when they appear and their previous (Qwen-2, Gemini-2.5) versions stay close (within $5-7$\,s), whereas Claude 4/4.5, GPT-5.2, and GPT-4.1 mini drift $20$--$65$\,s away. The programmatic timing metrics reveal pause-budget trade-offs among the strongest models. Gemini-3.5 Flash nearly eliminates overlap ($0.26$\,s) by having shorter descriptions (Coverage $0.45$) and achieves the shortest Avg.\ Silence ($4.85$\,s); Gemini-2.5 Flash balances higher silence use (Coverage $0.74$) with some overlap ($8.92$\,s; $6.63$\,s Avg.\ Silence); and the Qwen models fill more silence (Coverage $0.72$--$0.94$) at the cost of $30.45$--$52.14$\,s of overlaps.
No model leads on every metric; On-Screen Text, for example, stays within $0.56$--$0.64$ across all nine. %

\textbf{Holistic Script Quality.} Models scored with the Gemini-2.5 Flash judge (\cref{fig:model_comparison_videoally}) average $2.0$--$3.3$ on Descriptive, Objective, and Clear ($2.2$--$3.3$) yet only $1.16$--$1.63$ on Reference Alignment (\emph{Accurate}). The $\dagger$ models, scored by Gemini-3.5 Flash, score higher. Since the judge also changed, part of this jump may come from the judge rather than the generator. Low Reference Alignment partly reflects the reference-matching penalty described in \cref{sec:metrics}, which penalizes valid visual details omitted by the human describer and which \texttt{Quality-RF} resolves against human ratings in \cref{sec:human_studies}.

\begin{figure}[!bht]
\vspace{-0.4cm}
    \centering
        \includegraphics[width=0.35\columnwidth]{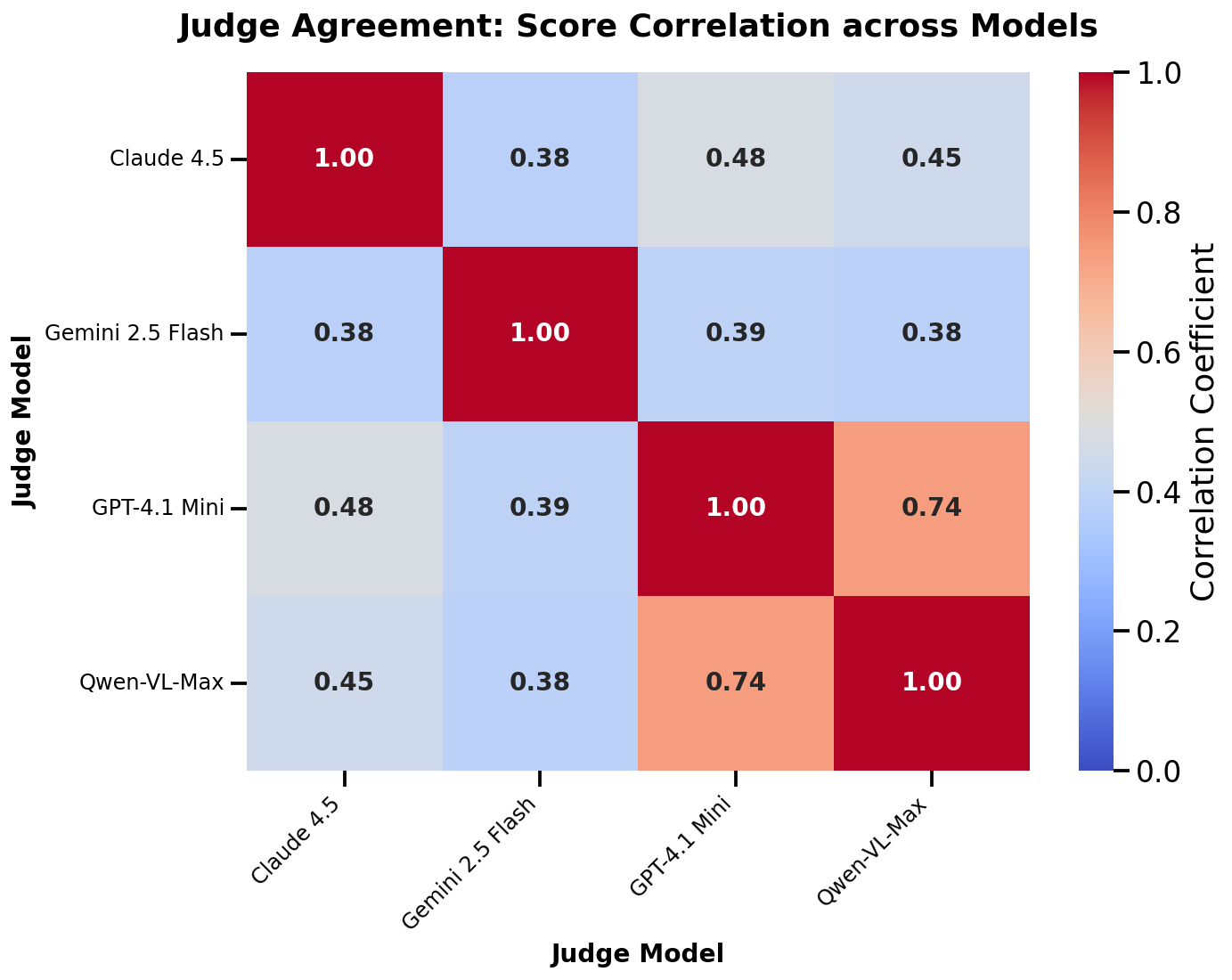}
        \includegraphics[width=0.33\columnwidth]{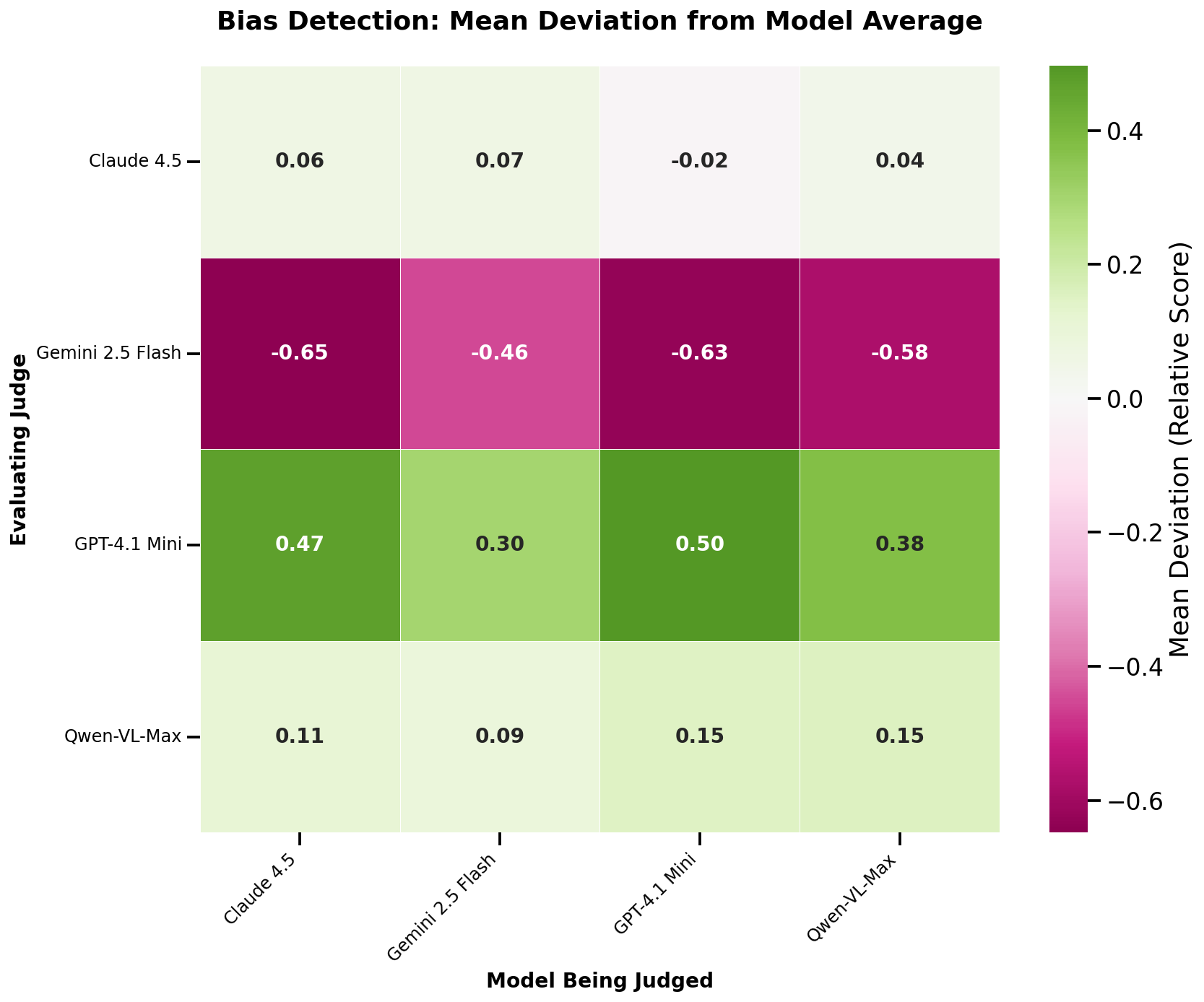}
    \caption{{\textbf{Judge Alignment and Bias.} 
    (a, left) Inter-judge Pearson correlation across 101 test videos. Each score is the mean across five metrics  Prioritization (G1.1) and \texttt{VideoA11y-Ref} rubrics; GPT-4.1 mini and Qwen-2-VL-Max align most closely. 
    (b, right) Deviation from model-specific mean scores. Shows systematic bias of each judge by calculating the mean deviation from the global average score per video. Positive values (green) indicate "lenient" judging, while negative values (pink) indicate "harsh" judging relative to the consensus.}}
    \label{fig:judge_analysis}
\end{figure}

\vspace{-0.2cm}
\subsection{LLM as Judge and Generalization}
\label{sec:llm_judge}
\vspace{-0.1cm}
\textbf{Inter-judge reliability.} As observed with judgments of script quality between Gemini 2.5 Flash and 3.5 Flash judges, unconstrained Likert scores shift with judge leniency, so some changes and jumps conflates generator gains with the judge change. To test sensitivity to the evaluator, we re-score the same ADs with Claude, Gemini, GPT and Qwen judges and correlate their per-video scores on five LLM-judged metrics (\cref{fig:judge_analysis}a) %
All judge pairs correlate positively ($r = 0.38$--$0.74$), and GPT-4.1 mini and Qwen-2-VL-Max agree most. However, calibration differs across judges (\cref{fig:judge_analysis}b): some judges are systematically more lenient (\textit{e.g.}, GPT-4.1 mini) and others harsher (\textit{e.g.}, Gemini-2.5-Flash), but these offsets are small. This supports using MLLM judges to rank systems, provided all systems in a comparison use the same judge.

\begin{figure*}[!tbh]
    \centering
        \includegraphics[width=0.74\textwidth]{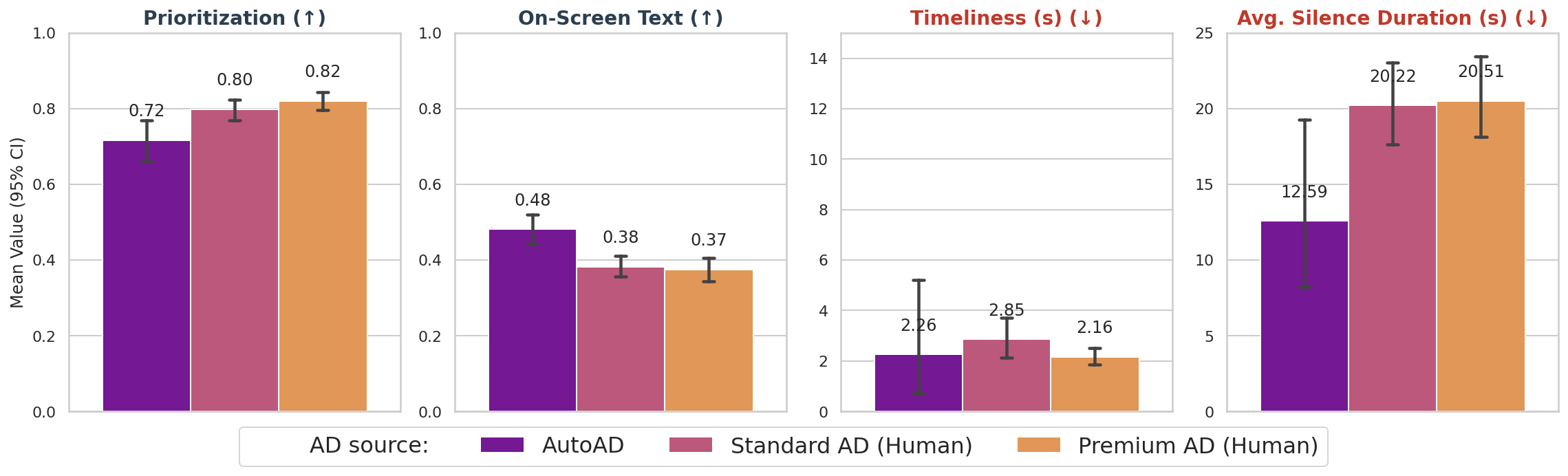}
        \includegraphics[width=0.24\textwidth]{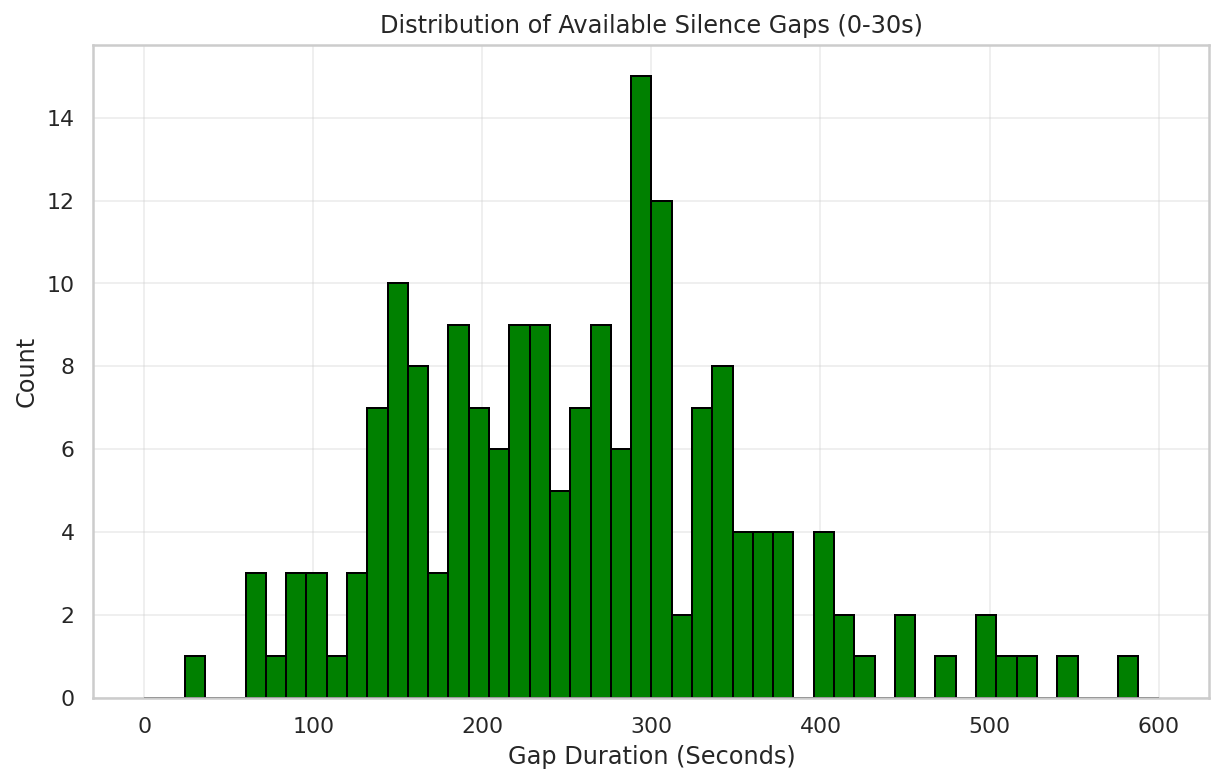}
    
    \caption{\textbf{YouDescribe generalization.} (a): Mean values (with 95\% CI) for Prioritization, On-Screen Text coverage, Timeliness, and Avg.\ Silence across 2 available human AD sources and our automated AD on the YouDescribe dataset. (b) Distribution of available silence gaps (right) shows  significantly longer gaps compared to our dataset due to the source videos being largely silent or lacking pre-existing dialogue.}
    \label{fig:youdescribe_analysis}
    \vspace{-0.4cm}
\end{figure*}

\textbf{Generalization to YouDescribe.} When applying our evaluation suite to external YouDescribe videos (\cref{fig:youdescribe_analysis}), we observe that ``Premium'' human AD trends ahead of ``Standard'' volunteer AD on Timeliness ($2.16$\,s vs.\ $2.85$\,s) and Prioritization ($0.82$ vs.\ $0.80$), though CIs overlap. This suggests that the metrics transfer to other user-generated videos and AD sources.

\vspace{-0.1cm}
\subsection{Sighted and BLV Participant Studies \& Metric Validation}
\label{sec:human_studies}
\vspace{-0.2cm}
We conducted complementary human studies with sighted raters and BLV participants using a custom playback interface. Sighted raters and BLV listeners serve distinct, complementary roles in evaluating AD~\cite{huh2024longform}: sighted raters can directly inspect the video frames to verify visual-to-text faithfulness, missing visual events, and exact visual-timestamp alignment across many model/prompt pairs, whereas BLV participants evaluate whether the audio track and descriptions form a coherent, non-intrusive, and engaging standalone narrative without visual context.

\begin{figure*}[bht]
\vspace{-0.3cm}
    \centering
    \includegraphics[width=\linewidth]{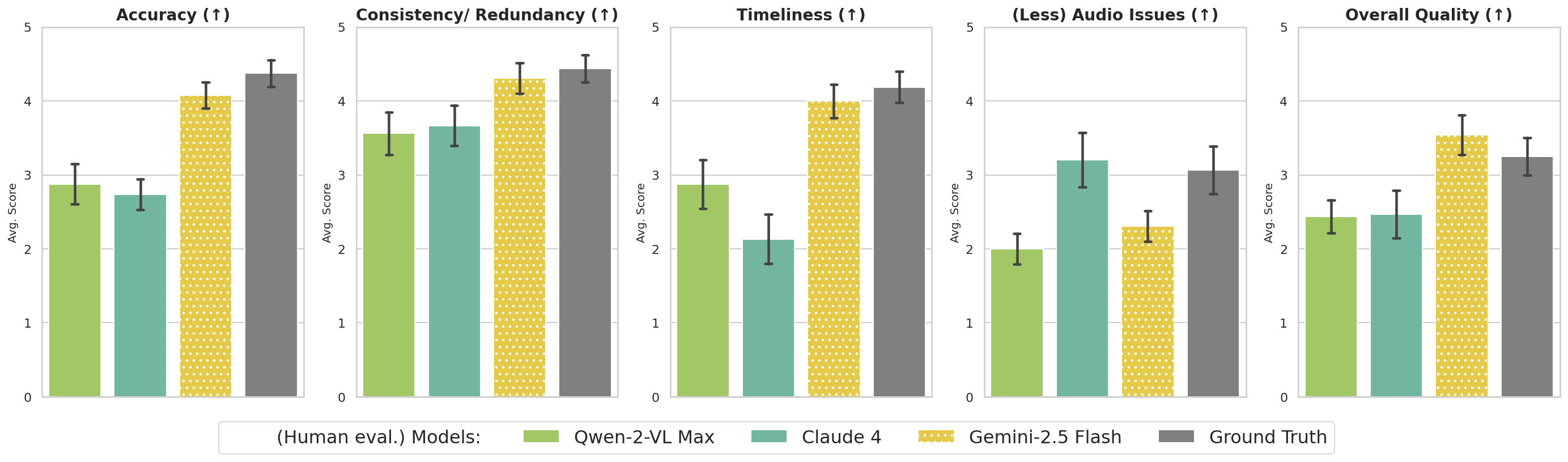}
    \caption{\textbf{Sighted rater model comparisons}. Mean Likert scores across five (of 11) key dimensions on videos evaluated by sighted raters ($\uparrow$ indicates higher is better). Gemini-2.5 Flash approaches Ground Truth on Accuracy, Timeliness and script Quality, while Claude 4 has fewer Audio Issues due to lower speech coverage.}
    \label{fig:sighted_model_comparison}
    \vspace{-0.2cm}
\end{figure*}

\textbf{Sighted Rater Study.} A pool of 8 sighted raters evaluated 42 videos, with AD instances spanning multiple models, prompts, and professional Human AD, across 11 dimensions including \texttt{Coverage}, \texttt{Accuracy}, \texttt{Relevance}, \texttt{Clarity}, \texttt{Timeliness}, and \texttt{Audio Issues} on a likert scale (full questionnaire in \cref{appx_sec:sighted_rater_qs}; and ratings of prompt comparison are in \cref{fig:sighted_prompt_comparison}).  Inter-annotator agreement (Krippendorff's $\alpha$, \cref{tab:inter_annotator_agreement_full}) across all metrics was $0.654$, led by Timeliness ($0.721$) and Audio Issues ($0.700$). Comparing models (\cref{fig:sighted_model_comparison}), Gemini-2.5 Flash with \textit{Gd.\,+ Timing ASR} prompt approaches human ground truth in Accuracy and Timeliness but lags on audio overlap issues. 

\begin{figure*}[t]
    \centering
    \vspace{-0.05cm}
    \includegraphics[width=0.8\linewidth]{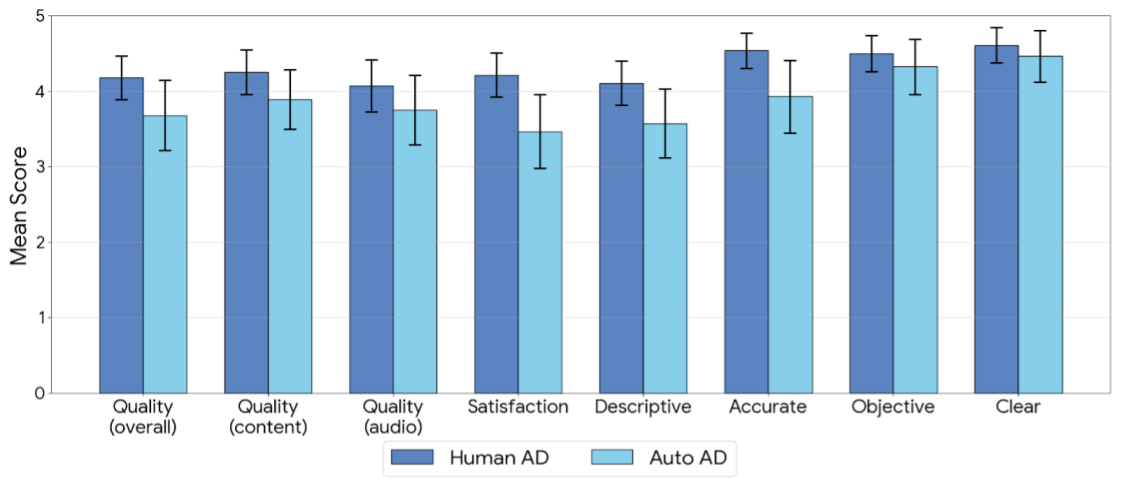}
    \caption{\textbf{BLV study}. Mean Likert scores across 8 key dimensions (error bars depict 95\% confidence intervals). BLV participants evaluated videos played with professional human written ADs (\mtgad) and AutoAD (Gemini-2.5 Flash with Gd. + Timing ASR). They rated both AD versions consistently high across all metrics and the human generated AD slightly consistently better in all dimensions. (Full likert scoring in \cref{fig:blv_pref})}
    \label{fig:blv_ratings}
    \vspace{-0.3cm}
\end{figure*}

\vspace{-0.2cm}
\textbf{BLV Participant Study.} We recruited 7 BLV participants who regularly watch YouTube. Each participant evaluated four videos randomly sampled from our dataset ($n = 28$ video-level evaluations), listening to both AutoAD (Gemini-2.5 Flash with \textit{Gd. + Timing ASR}) and the professionally authored Human AD. Participants were unaware of the AD source and we randomized and counterbalanced the order of the AD sources. Participants provided ratings for each AD (Fig.~\ref{fig:blv_ratings}), their preference between the two AD (Appx. Fig.~\ref{fig:blv_pref}), and qualitative feedback about each AD. 
Participants rated both versions consistently high across all dimensions (\cref{fig:blv_ratings}). While participants marginally preferred human-authored AD ($53.6\%$, \cref{fig:blv_pref}), Wilcoxon Signed-Rank tests showed statistically significant differences only for overall Satisfaction ($p < 0.05$) and perceived Accuracy ($p < 0.05$), with near-parity on Quality, Clarity, and Objectivity. Individual preferences varied markedly across content genres (\cref{appx_sec:blv_qualitative}): several participants preferred the denser visual detail of AutoAD for action- or travel-oriented clips while favoring the concise narrative pacing of professional human AD for dialogue-heavy vlogs.

\begin{table*}[!t]
\centering
\caption{\textbf{Empirical validation of automated metrics against Sighted ($n=42$) and BLV ($n=28$) human judgments.} Spearman rank correlations ($\rho$) and $p$-values across our Reference-Free Guideline Metrics (\textbf{G1--G8}), Reference-Based \texttt{VideoA11y-Ref}, and Reference-Free \texttt{Quality-RF} (\texttt{video + ASR}). Full qualitative interpretations and 4-variant ablations are in \cref{appx_sec:metric_validation}.}
\label{tab:human_metric_correlations}
\resizebox{\textwidth}{!}{%
\begin{tabular}{@{}llllrr@{}}
\toprule
\textbf{Category} & \textbf{Automated Metric} & \textbf{Cohort} & \textbf{Target Human Dimension} & \textbf{Spearman $\rho$} & \textbf{$p$-value} \\
\midrule
\multirow{6}{*}{\shortstack[l]{\textbf{Reference-Free}\\\textbf{Guideline Metrics}\\\textbf{(G1, G2, G7, G8)}}}
 & \textbf{Audio Overlaps (G7)} & Sighted ($n=42$) & Audio Issues (No Overlaps) & $\mathbf{-0.467}$ & $\mathbf{0.0018^{**}}$ \\
 & \textbf{Avg.\ Silence (G8)} & Sighted ($n=42$) & Gaps of Silence (Few Quiet Gaps) & $\mathbf{-0.415}$ & $\mathbf{0.0063^{**}}$ \\
 & \textbf{Timeliness (G2)} & Sighted ($n=42$) & Timeliness & $-0.228$ & $0.1470$ \\
 & \textbf{Prioritization (G1.1)} & Sighted ($n=42$) & Coverage & $+0.246$ & $0.1160$ \\
 & \textbf{No Redundancy (G1.2)} & BLV ($n=28$) & Consistency / Non-Redundancy & $\mathbf{+0.386}$ & $\mathbf{0.0427^{*}}$ \\
 & \textbf{Audio Overlaps (G7)} & BLV ($n=28$) & Audio Quality (truncated in UI) & $+0.132$ & $0.5024$ \\
\midrule
\multirow{6}{*}{\shortstack[l]{\textbf{Reference-Based}\\\textbf{VideoA11y-Ref}\\\cite{li2025videoa11y}}}
 & \textbf{VideoA11y-Ref Clear} & Sighted ($n=42$) & Clarity & $\mathbf{+0.453}$ & $\mathbf{0.0026^{**}}$ \\
 & \textbf{VideoA11y-Ref Descriptive} & Sighted ($n=42$) & Descriptive & $\mathbf{+0.328}$ & $\mathbf{0.0338^{*}}$ \\
 & \textbf{VideoA11y-Ref Clear} & BLV ($n=28$) & Clarity & $\mathbf{+0.388}$ & $\mathbf{0.0411^{*}}$ \\
 & \textbf{VideoA11y-Ref Descriptive} & BLV ($n=28$) & Descriptive & $+0.340$ & $0.0771$ \\
 & \textbf{VideoA11y-Ref Objective} & BLV ($n=28$) & Objective & $+0.335$ & $0.0811$ \\
 & \textbf{VideoA11y-Ref Accurate} & BLV ($n=28$) & Perceived Accuracy & $-0.264$ & $0.1743$ \\
\midrule
\multirow{4}{*}{\shortstack[l]{\textbf{Reference-Free}\\\textbf{Quality-RF}\\\textbf{(Video + ASR, Ours)}}}
 & \textbf{Quality-RF Descriptive} & BLV ($n=28$) & Descriptive & $\mathbf{+0.583}$ & $\mathbf{0.0011^{**}}$ \\
 & \textbf{Quality-RF Clear} & BLV ($n=28$) & Clarity & $\mathbf{+0.415}$ & $\mathbf{0.0281^{*}}$ \\
 & \textbf{Quality-RF Objective} & BLV ($n=28$) & Objective & $+0.365$ & $0.0561$ \\
 & \textbf{Quality-RF Accurate} & BLV ($n=28$) & Perceived Accuracy & $+0.244$ & $0.2109$ \\
\bottomrule
\end{tabular}%
}
\vspace{-0.3cm}
\end{table*}

\textbf{Do the automated metrics agree with people?} We compute video-level Spearman rank correlations between each metric and the matching human rating for both cohorts (\cref{tab:human_metric_correlations}; full results in \cref{appx_sec:metric_validation}). For sighted raters, each data point is a rated video with its AD. For BLV listeners, each data point is one participant's rating of the AutoAD version of a video ($n=28$), paired with that AD's metric scores. Negative $\rho$ is expected for metrics measured in seconds, since fewer seconds of overlap or silence should mean a higher 1--5 rating. For sighted raters, the two programmatic timing metrics are the strongest predictors: Audio Overlaps (G7) tracks ratings of audio issues and Avg.\ Silence (G8) tracks ratings of quiet gaps (both $p<0.01$). \texttt{VideoA11y-Ref} Clear and Descriptive also correlate significantly with the matching ratings. For BLV listeners, No Redundancy (G1.2) and \texttt{VideoA11y-Ref} Clear are significant predictors. Audio Overlaps correlates only weakly with BLV ratings of audio quality, by design of the study: our video player cut off any AD segment as soon as speech began. BLV listeners therefore never heard AD collide with speech; AD  that was too long instead showed up as abrupt cutoffs, which listeners penalized under Clarity.

\textbf{Resolving reference-matching bias with \texttt{Quality-RF}.}  \texttt{VideoA11y-Ref} Accurate correlates \emph{negatively} with BLV ratings of accuracy ($\rho=-0.264$). Two factors explain this. First, listeners without visual access judge accuracy by whether the narration is plausible and consistent~\cite{huh2024longform}. Second, the reference-based score penalizes correct details that the human describer omitted. Scoring against the video and transcript with \texttt{Quality-RF} eliminates this penalty, flipping Accuracy positive ($\rho=+0.244$) and substantially strengthens agreement on Descriptive ($\rho=+0.583$, $p=0.001$) and Clear ($\rho=+0.415$, $p=0.028$). Together, the guideline metrics and \texttt{Quality-RF} form a fully reference-free evaluation suite that is validated against human judgments and works on unseen videos.

\vspace{-0.2cm}
\section{Conclusion and Broader Impact}
\vspace{-0.2cm}
While frontier MLLMs approach human AD in visual descriptiveness, there's still a clear gap in temporal precision, remaining silence, and narrative pacing for inline AD. Through the \mtgad dataset with 36 hours of high-quality ground truth, co-created by teams of blind and sighted professionals, and \mtgadeval, an end-to-end reference-free evaluation suite, we demonstrate: 
\begin{itemize}[nosep, leftmargin=*]
    \item \textbf{Evaluation-Guided Prompting for Precision:} The framework allows development of prompts and harnesses for improved audio description generation. We find that incorporating ASR transcripts and temporal reasoning (via the \textit{Gd. + Timing ASR} strategy) is essential for balancing content richness with strict temporal constraints, reducing audio overlaps.
    \item \textbf{Model Performance Benchmarking:} Newer-generation omni-modal models (Gemini-3.5 Flash, Qwen-3.5-Omni) raise content prioritization from $0.76$ to $0.79$--$0.94$, and Gemini-3.5 Flash cuts speech overlaps to $0.26$\,s. Still, no model leads on every metric: the models trade overlap against coverage of the silences, and human AD remains better paced.
    \item \textbf{User Utility and Reference-Free Evaluation:} Our study with blind and low-vision (BLV) participants confirms that while human-crafted AD remains the gold standard for narrative flow, MLLM-generated AD is already highly beneficial for comprehension, and \mtgadeval, our validated reference-free metric suite \mtgadeval (\textbf{G1.1--G1.3, G2, G7, G8} and \texttt{Quality-RF}) correlates substantially with both sighted and BLV human judgments.
\end{itemize}

\textbf{Guidance for System Designers.} Inline AD is a multi-objective constrained problem, and a single aggregate score hides failure modes (\textit{e.g.}, high Prioritization with heavy overlaps). For model selection we recommend two stages: (1) \textbf{Temporal Feasibility Filtering}, a hard threshold on \textit{Audio Overlaps (G7)} so AD does not mask dialogue, and \textit{Audio Coverage (G8)} $\le 1$ to rule out over-generation (this still leaves some gaps undescribed, as Avg.\ Silence shows); then (2) \textbf{Content \& Narrative Optimization}, maximizing \textit{Prioritization (G1.1)}, \textit{No Redundancy (G1.2)}, and \textit{Reference-Free Clarity/Descriptiveness (\texttt{Quality-RF})} among temporally feasible candidates.

\textbf{Limitations.}
Our automated metrics cover six guidelines from our BLV pilot (\textbf{G1.1--G1.3, G2, G7, G8}), leaving others such as active voice (\textbf{G5}) and prosody (\textbf{G10}) to future work; G1.1 and G2 also depend on MLLM-extracted events ($\mathcal{D}$) and judge calibration, which can under-credit subtle human-observed details. Second, our BLV study ($7$ participants, $28$ full-length video evaluations) gives rich paired comparisons, but larger cohorts across more genres would better capture individual preferences. Finally, our temporal metrics render AD with a standard TTS engine at a fixed $1.2\times$ speaking rate, and absolute overlap and coverage values depend on this rate. We leave the speaking rate to future work; faster or adaptive per-gap speech could fit richer descriptions into the silences.

\textbf{Dataset and Code Release.} We release \mtgad under CC-BY-4.0 (\mtgdataurl) and \mtgadeval under Apache-2.0 (\mtgcodeurl). Bundling human and model ADs, ASR transcripts, and pre-extracted visual events ($\mathcal{D}$) to support reproducibility, re-use and further development of the evaluation frameworks.

\textbf{Broader Impact.}
The \mtgad dataset and \mtgadeval reference-free evaluation framework can support automated auditing and generation of inline AD for user-generated tutorials, educational content, and vlogs at scale, and could serve as video accessibility indicators for BLV audiences. It also carries risks: generated AD can hallucinate details that mislead BLV users or misrepresent the creator's intent, which listeners cannot easily verify, so AI-generated AD should be labeled as such.

\clearpage
\section*{Acknowledgements}
We thank Cindy Bennett and Shaun Kane for their valuable feedback on the audio description guidelines, and Aidan Bryant for help with an early BLV pilot survey. We also thank Philip Nelson and Chandralekha Motati for their support of this work, as well as the audio description professionals, sighted raters, and blind and low-vision study participants for their annotations and insights.

\bibliographystyle{plain}
\bibliography{main}

\newpage

\appendix
\crefalias{section}{appsection}
\crefalias{figure}{appfigure}
\crefalias{table}{apptable}

\section*{Appendix}

\section{Synthesized Suite of metrics from Expert Guidelines}
\label{appx_sec:full_eval_metrics}

This section presents the full suite of evaluation metrics we synthesized based on expert guidelines.

\noindent \textbf{AD Script Content Quality: }
\begin{itemize}
    \item \textbf{G1 - Content Coverage:} AD describes important visual content necessary to understand the video.
    \begin{itemize}
        \item \textbf{G1.1 - Prioritization:} AD prioritizes visual information necessary to understand the video. 
        \item \textbf{G1.2 - Redundancy:} AD does not describe information that is already clear to BLV viewers from the audio alone (\textit{e.g.}, redundant with speech or sounds).
        \item \textbf{G1.3 - On-Screen Text:} AD narrates on-screen text.
        \item \textbf{G1.4 - Visual Style:} AD describes style, color, textures and temporal transitions only as useful to understand the video (audience members differ in their preferences for descriptions around visual styles~\cite{liu2021what,jiang2024s,natalie2024audio}). 
    \end{itemize}
    \item \textbf{G2 - Content Timeliness:} AD describes visual content close to the time it appears on screen. 
    \item \textbf{G3 - Content Accuracy:} AD contains only accurate information (similar metrics already addressed in prior work~\cite{han2024autoad}).
    \item \textbf{G4 - Content Objectivity:} AD describes content as it appears on screen and does not include censorship or interpretation (audience members differ in their preferences for objective vs. subjective descriptions~\cite{liu2021what,jiang2024s}). 
\end{itemize}

\noindent \textbf{AD Script Style: }
\begin{itemize}
    \item \textbf{G5 - Active Voice:} AD script is written in third person with active voice. 
    \item \textbf{G6 - Appropriateness:} Language of the AD matches the language of the video such that it is appropriate for the audience.
    \begin{itemize}
        \item \textbf{G6.1 - Consistency:} AD uses language that is consistent within the AD and with the speech in the video. 
        \item \textbf{G6.2 - Terminology and Tone:} AD uses vocabulary and tone that are similar to the vocabulary and tone in the video (\textit{e.g.}, does not introduce jargon, uses similar level of formality).
    \end{itemize}
\end{itemize}

\noindent \textbf{AD Audio Quality: }
\begin{itemize}
    \item \textbf{G7 - Overlaps}: AD avoids overlapping speech.
    \item \textbf{G8 - Audio Coverage}: AD covers the majority of the video, but does not need to fill every silence. 
    \item \textbf{G9 - Pronunciation}: AD pronounces words correctly.
    \item \textbf{G10 - Appropriate Speech Style: } Speed and prosody are appropriate for the content of the video (\textit{e.g.}, somber for a melancholy scene). 
\end{itemize}

For the analysis of the main experiments in this work, we use a subset of our guidelines that we prioritized for evaluation based on a pilot study with BLV audience members reviewing descriptions created with a basic prompt (\textit{e.g.}, ``Create an audio description''). In the pilot study BLV users surfaced issues that most negatively impacted their experience, and these issues informed our choice of guidelines for initial metrics (\textbf{G1.1-1.3}, \textbf{G2}, \textbf{G7}, \textbf{G8}).

\section{Full Results on the Proposed Metrics}
\label{appx_sec:full_results}
This section gives the full results behind \cref{fig:prompt_comparison_main,fig:model_comparison_main,fig:model_comparison_videoally}: the prompt comparison (\cref{tab:main_results}), the six initial models (\cref{tab:main_res_model_comp}), and the newer omni-modal models (\cref{tab:omninew_models_comp}). The prompt comparison uses the validation videos for which every prompt has output. The model tables score each model on all 101 test videos it has output for (Qwen-3.5-Omni-Flash: 57, Qwen-3.5-Omni-Plus: 99), so a model's values agree across tables, but Gemini-2.5 Flash with the Gd.\ + Timing ASR prompt differs between the prompt table and the model tables. Audio Coverage is better when higher only up to $1$; values above $1$ mean the AD is longer than all available silence (\cref{sec:metrics}). We therefore shade the highest values that are $\le 1$.

\begin{table}[h]
\centering
\resizebox{\columnwidth}{!}
{
\begin{tabular}{@{}lccccc@{}}
\toprule
\textbf{Metric} & \textbf{Simple AD} & \textbf{Guideline only} & \textbf{Guideline + ASR} & \textbf{Gd. + Timing ASR} & \textbf{Human AD} \\
\midrule
Prioritization, G1.1 ($\uparrow$) & 0.75 \tiny{$\pm$0.02} & \cellcolor[HTML]{E2E2E2} 0.79 \tiny{$\pm$0.02} & \cellcolor[HTML]{F2F2F2} 0.79 \tiny{$\pm$0.02} & 0.76 \tiny{$\pm$0.02} & 0.73 \tiny{$\pm$0.02} \\
No Redundancy, G1.2 ($\uparrow$) & \cellcolor[HTML]{E2E2E2} 0.77 \tiny{$\pm$0.03} & 0.72 \tiny{$\pm$0.02} & 0.71 \tiny{$\pm$0.02} & 0.69 \tiny{$\pm$0.03} & \cellcolor[HTML]{F2F2F2} 0.72 \tiny{$\pm$0.02} \\
On-Screen Text, G1.3 ($\uparrow$) & 0.47 \tiny{$\pm$0.03} & \cellcolor[HTML]{E2E2E2} 0.69 \tiny{$\pm$0.03} & \cellcolor[HTML]{F2F2F2} 0.65 \tiny{$\pm$0.03} & 0.63 \tiny{$\pm$0.03} & 0.61 \tiny{$\pm$0.03} \\
Timeliness, G2 (s) ($\downarrow$) & 6.46 \tiny{$\pm$3.03} & \cellcolor[HTML]{F2F2F2} 3.27 \tiny{$\pm$1.29} & 5.11 \tiny{$\pm$2.07} & 4.43 \tiny{$\pm$1.17} & \cellcolor[HTML]{E2E2E2} 2.90 \tiny{$\pm$0.48} \\
Audio Overlaps, G7 (s) ($\downarrow$) & 15.07 \tiny{$\pm$3.01} & 59.19 \tiny{$\pm$7.28} & 21.34 \tiny{$\pm$4.64} & \cellcolor[HTML]{F2F2F2} 12.77 \tiny{$\pm$3.88} & \cellcolor[HTML]{E2E2E2} 7.51 \tiny{$\pm$1.42} \\
Audio Coverage, G8 (ratio, $\le$1) & 0.41 \tiny{$\pm$0.06} & 1.23 \tiny{$\pm$0.11} & \cellcolor[HTML]{E2E2E2} 0.83 \tiny{$\pm$0.08} & \cellcolor[HTML]{F2F2F2} 0.78 \tiny{$\pm$0.05} & 0.61 \tiny{$\pm$0.03} \\
Avg.\ Silence, G8 (s) ($\downarrow$) & 17.95 \tiny{$\pm$3.46} & 8.85 \tiny{$\pm$3.13} & \cellcolor[HTML]{F2F2F2} 6.71 \tiny{$\pm$2.47} & 8.97 \tiny{$\pm$2.80} & \cellcolor[HTML]{E2E2E2} 4.19 \tiny{$\pm$1.13} \\ \midrule
VideoA11y-Ref Descriptive ($\uparrow$) & \cellcolor[HTML]{F2F2F2} 3.46 \tiny{$\pm$0.13} & 3.28 \tiny{$\pm$0.14} & 2.81 \tiny{$\pm$0.11} & 3.03 \tiny{$\pm$0.10} & \cellcolor[HTML]{E2E2E2} 4.72 \tiny{$\pm$0.07} \\
VideoA11y-Ref Objective ($\uparrow$) & 3.55 \tiny{$\pm$0.13} & \cellcolor[HTML]{F2F2F2} 3.59 \tiny{$\pm$0.13} & 3.27 \tiny{$\pm$0.12} & 2.98 \tiny{$\pm$0.13} & \cellcolor[HTML]{E2E2E2} 4.93 \tiny{$\pm$0.03} \\
VideoA11y-Ref Accurate ($\uparrow$) & 1.82 \tiny{$\pm$0.11} & \cellcolor[HTML]{F2F2F2} 2.18 \tiny{$\pm$0.12} & 1.74 \tiny{$\pm$0.10} & 1.64 \tiny{$\pm$0.09} & \cellcolor[HTML]{E2E2E2} 4.98 \tiny{$\pm$0.02} \\
VideoA11y-Ref Clear ($\uparrow$) & 2.59 \tiny{$\pm$0.13} & \cellcolor[HTML]{F2F2F2} 2.85 \tiny{$\pm$0.14} & 2.57 \tiny{$\pm$0.12} & 2.68 \tiny{$\pm$0.11} & \cellcolor[HTML]{E2E2E2} 4.96 \tiny{$\pm$0.03} \\
\bottomrule
\end{tabular}
}
\caption{\textbf{Prompt Comparisons}. AD generated by Gemini-2.5 Flash with each prompt, and the professional Human AD, on validation videos with output from every prompt. Showing the mean and the 95\% confidence interval, $\uparrow$ (higher is better); $\downarrow$ (lower is better). Cells are shaded by rank: \colorbox[HTML]{E2E2E2}{First} and \colorbox[HTML]{F2F2F2}{Second}.}
\label{tab:main_results}
\end{table}

\begin{table*}[h]
\centering
\resizebox{\columnwidth}{!}{
\begin{tabular}{@{}lcccccc@{}}
\toprule
\textbf{Metric} & \textbf{Gemini-2.5 Flash} & \textbf{Qwen-2-VL-Max} & \textbf{GPT-4.1 mini} & \textbf{Claude 4} & \textbf{Claude 4.5} & \textbf{GPT-5.2} \\
\midrule
Prioritization, G1.1 ($\uparrow$) & \cellcolor[HTML]{E2E2E2} 0.76 \tiny{$\pm$0.04} & \cellcolor[HTML]{F2F2F2} 0.59 \tiny{$\pm$0.05} & 0.52 \tiny{$\pm$0.05} & 0.51 \tiny{$\pm$0.05} & \cellcolor[HTML]{F2F2F2} 0.59 \tiny{$\pm$0.05} & 0.55 \tiny{$\pm$0.05} \\
No Redundancy, G1.2 ($\uparrow$) & 0.70 \tiny{$\pm$0.05} & \cellcolor[HTML]{F2F2F2} 0.71 \tiny{$\pm$0.04} & 0.70 \tiny{$\pm$0.05} & 0.69 \tiny{$\pm$0.05} & 0.70 \tiny{$\pm$0.04} & \cellcolor[HTML]{E2E2E2} 0.73 \tiny{$\pm$0.04} \\
On-Screen Text, G1.3 ($\uparrow$) & \cellcolor[HTML]{E2E2E2} 0.64 \tiny{$\pm$0.06} & 0.59 \tiny{$\pm$0.06} & 0.58 \tiny{$\pm$0.06} & 0.56 \tiny{$\pm$0.05} & 0.61 \tiny{$\pm$0.05} & \cellcolor[HTML]{F2F2F2} 0.63 \tiny{$\pm$0.05} \\
Timeliness, G2 (s) ($\downarrow$) & \cellcolor[HTML]{E2E2E2} 5.00 \tiny{$\pm$3.07} & \cellcolor[HTML]{F2F2F2} 6.92 \tiny{$\pm$3.91} & 65.28 \tiny{$\pm$18.59} & 19.83 \tiny{$\pm$4.39} & 20.50 \tiny{$\pm$4.30} & 22.56 \tiny{$\pm$4.76} \\
Audio Overlaps, G7 (s) ($\downarrow$) & \cellcolor[HTML]{F2F2F2} 8.92 \tiny{$\pm$5.62} & 46.53 \tiny{$\pm$11.16} & 21.46 \tiny{$\pm$4.77} & \cellcolor[HTML]{E2E2E2} 7.45 \tiny{$\pm$2.14} & 12.86 \tiny{$\pm$3.31} & 11.14 \tiny{$\pm$2.74} \\
Audio Coverage, G8 (ratio, $\le$1) & \cellcolor[HTML]{F2F2F2} 0.74 \tiny{$\pm$0.07} & \cellcolor[HTML]{E2E2E2} 0.94 \tiny{$\pm$0.15} & 1.48 \tiny{$\pm$0.46} & 0.62 \tiny{$\pm$0.06} & 0.70 \tiny{$\pm$0.07} & 0.65 \tiny{$\pm$0.07} \\
Avg.\ Silence, G8 (s) ($\downarrow$) & \cellcolor[HTML]{E2E2E2} 6.63 \tiny{$\pm$2.16} & \cellcolor[HTML]{F2F2F2} 8.81 \tiny{$\pm$2.78} & 19.37 \tiny{$\pm$5.89} & 18.57 \tiny{$\pm$8.78} & 10.40 \tiny{$\pm$5.86} & 14.85 \tiny{$\pm$7.31} \\ \midrule
VideoA11y-Ref Descriptive ($\uparrow$) & \cellcolor[HTML]{F2F2F2} 3.10 \tiny{$\pm$0.19} & 2.58 \tiny{$\pm$0.22} & \cellcolor[HTML]{E2E2E2} 3.28 \tiny{$\pm$0.24} & 2.32 \tiny{$\pm$0.18} & 2.69 \tiny{$\pm$0.21} & 2.89 \tiny{$\pm$0.21} \\
VideoA11y-Ref Objective ($\uparrow$) & \cellcolor[HTML]{F2F2F2} 2.92 \tiny{$\pm$0.22} & 2.77 \tiny{$\pm$0.25} & 2.82 \tiny{$\pm$0.22} & 2.13 \tiny{$\pm$0.20} & 2.77 \tiny{$\pm$0.24} & \cellcolor[HTML]{E2E2E2} 2.94 \tiny{$\pm$0.26} \\
VideoA11y-Ref Accurate ($\uparrow$) & \cellcolor[HTML]{E2E2E2} 1.63 \tiny{$\pm$0.15} & 1.33 \tiny{$\pm$0.14} & 1.31 \tiny{$\pm$0.14} & 1.16 \tiny{$\pm$0.10} & 1.26 \tiny{$\pm$0.13} & \cellcolor[HTML]{F2F2F2} 1.35 \tiny{$\pm$0.14} \\
VideoA11y-Ref Clear ($\uparrow$) & \cellcolor[HTML]{E2E2E2} 2.69 \tiny{$\pm$0.20} & 2.21 \tiny{$\pm$0.21} & 2.19 \tiny{$\pm$0.23} & 2.01 \tiny{$\pm$0.18} & 2.21 \tiny{$\pm$0.22} & \cellcolor[HTML]{F2F2F2} 2.54 \tiny{$\pm$0.25} \\
\bottomrule
\end{tabular}
}
\caption{\textbf{Model Comparisons (Initial Frontier Models)}. Comparing frontier models under the \textit{Gd. + Timing ASR} prompt on the 101 test videos. Showing the mean and the 95\% confidence interval, $\uparrow$ (higher is better); $\downarrow$ (lower is better). Cells are shaded by rank: \colorbox[HTML]{E2E2E2}{First} and \colorbox[HTML]{F2F2F2}{Second}.}
\label{tab:main_res_model_comp}
\end{table*}

\begin{table*}[h]
\centering
\resizebox{\columnwidth}{!}{
\begin{tabular}{@{}lccccc@{}}
\toprule
\textbf{Metric} & \textbf{Gemini-2.5 Flash} & \textbf{Gemini-3.5 Flash$^{\dagger}$} & \textbf{Qwen-2-VL-Max} & \textbf{Qwen-3.5-Omni-Flash$^{\dagger}$} & \textbf{Qwen-3.5-Omni-Plus$^{\dagger}$} \\
\midrule
Prioritization, G1.1 ($\uparrow$) & 0.76 \tiny{$\pm$0.04} & \cellcolor[HTML]{F2F2F2} 0.93 \tiny{$\pm$0.03} & 0.59 \tiny{$\pm$0.05} & \cellcolor[HTML]{E2E2E2} 0.94 \tiny{$\pm$0.02} & 0.79 \tiny{$\pm$0.06} \\
No Redundancy, G1.2 ($\uparrow$) & 0.70 \tiny{$\pm$0.05} & \cellcolor[HTML]{E2E2E2} 0.90 \tiny{$\pm$0.03} & 0.71 \tiny{$\pm$0.04} & \cellcolor[HTML]{F2F2F2} 0.88 \tiny{$\pm$0.05} & \cellcolor[HTML]{F2F2F2} 0.88 \tiny{$\pm$0.04} \\
On-Screen Text, G1.3 ($\uparrow$) & \cellcolor[HTML]{E2E2E2} 0.64 \tiny{$\pm$0.06} & 0.57 \tiny{$\pm$0.05} & 0.59 \tiny{$\pm$0.06} & \cellcolor[HTML]{F2F2F2} 0.61 \tiny{$\pm$0.06} & 0.58 \tiny{$\pm$0.06} \\
Timeliness, G2 (s) ($\downarrow$) & 5.00 \tiny{$\pm$3.07} & \cellcolor[HTML]{F2F2F2} 2.99 \tiny{$\pm$0.91} & 6.92 \tiny{$\pm$3.91} & 4.91 \tiny{$\pm$3.64} & \cellcolor[HTML]{E2E2E2} 2.67 \tiny{$\pm$1.04} \\
Audio Overlaps, G7 (s) ($\downarrow$) & \cellcolor[HTML]{F2F2F2} 8.92 \tiny{$\pm$5.62} & \cellcolor[HTML]{E2E2E2} 0.26 \tiny{$\pm$0.13} & 46.53 \tiny{$\pm$11.16} & 52.14 \tiny{$\pm$15.89} & 30.45 \tiny{$\pm$8.69} \\
Audio Coverage, G8 (ratio, $\le$1) & \cellcolor[HTML]{F2F2F2} 0.74 \tiny{$\pm$0.07} & 0.45 \tiny{$\pm$0.03} & \cellcolor[HTML]{E2E2E2} 0.94 \tiny{$\pm$0.15} & \cellcolor[HTML]{E2E2E2} 0.94 \tiny{$\pm$0.20} & 0.72 \tiny{$\pm$0.14} \\
Avg.\ Silence, G8 (s) ($\downarrow$) & \cellcolor[HTML]{F2F2F2} 6.63 \tiny{$\pm$2.16} & \cellcolor[HTML]{E2E2E2} 4.85 \tiny{$\pm$1.20} & 8.81 \tiny{$\pm$2.78} & 16.61 \tiny{$\pm$11.88} & 9.49 \tiny{$\pm$2.12} \\ \midrule
VideoA11y-Ref Descriptive ($\uparrow$) & 3.10 \tiny{$\pm$0.19} & \cellcolor[HTML]{F2F2F2} 4.02 \tiny{$\pm$0.15} & 2.58 \tiny{$\pm$0.22} & \cellcolor[HTML]{E2E2E2} 4.43 \tiny{$\pm$0.23} & 3.82 \tiny{$\pm$0.27} \\
VideoA11y-Ref Objective ($\uparrow$) & 2.92 \tiny{$\pm$0.22} & \cellcolor[HTML]{F2F2F2} 4.07 \tiny{$\pm$0.15} & 2.77 \tiny{$\pm$0.25} & \cellcolor[HTML]{E2E2E2} 4.41 \tiny{$\pm$0.20} & 4.03 \tiny{$\pm$0.26} \\
VideoA11y-Ref Accurate ($\uparrow$) & 1.63 \tiny{$\pm$0.15} & \cellcolor[HTML]{F2F2F2} 3.79 \tiny{$\pm$0.18} & 1.33 \tiny{$\pm$0.14} & \cellcolor[HTML]{E2E2E2} 3.98 \tiny{$\pm$0.25} & 3.23 \tiny{$\pm$0.25} \\
VideoA11y-Ref Clear ($\uparrow$) & 2.69 \tiny{$\pm$0.20} & \cellcolor[HTML]{F2F2F2} 4.02 \tiny{$\pm$0.17} & 2.21 \tiny{$\pm$0.21} & \cellcolor[HTML]{E2E2E2} 4.27 \tiny{$\pm$0.26} & 3.81 \tiny{$\pm$0.27} \\
\bottomrule
\end{tabular}
}
\caption{\textbf{Next-Generation Omni-Modal Model Comparisons}. Comparing Gemini-2.5/3.5 Flash and Qwen-2-VL-Max vs.\ Qwen-3.5-Omni-Flash/Plus under the \textit{Gd. + Timing ASR} prompt. Each model is scored on all test videos it has output for: 101 for Gemini-2.5/3.5 Flash and Qwen-2-VL-Max, 99 for Qwen-3.5-Omni-Plus, and 57 for Qwen-3.5-Omni-Flash. Showing the mean and 95\% confidence interval, $\uparrow$ (higher is better); $\downarrow$ (lower is better). For models marked with $^{\dagger}$, all LLM-judged metrics (G1.1, G1.2, G2, and \texttt{VideoA11y-Ref}) used Gemini-3.5 Flash as the judge; the other models used Gemini-2.5 Flash. The segment-level metrics (G1.1, G1.2, G2) appear robust to this change, whereas the holistic \texttt{VideoA11y-Ref} ratings are more sensitive to it; programmatic metrics (G1.3, G7, G8) do not depend on the judge. Cells are shaded by rank: \colorbox[HTML]{E2E2E2}{First} and \colorbox[HTML]{F2F2F2}{Second}.}
\label{tab:omninew_models_comp}
\end{table*}

\newpage
\section{Inter Annotator Agreement and Sighted Rater Prompt Comparisons}
\begin{figure*}[!ht]
    \centering
    \includegraphics[width=\linewidth]{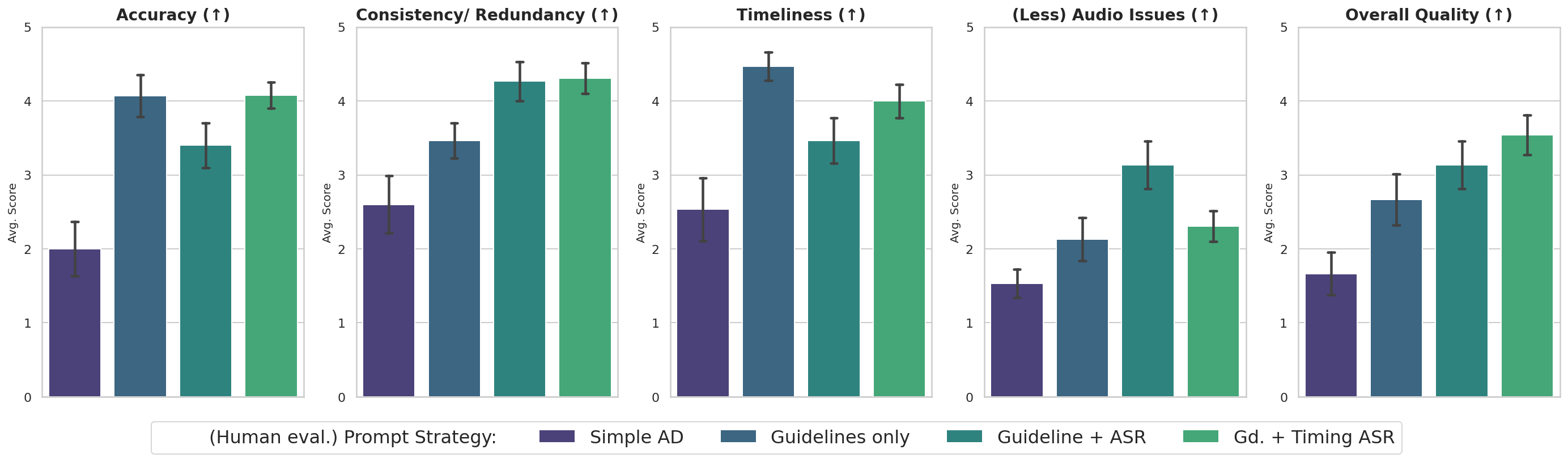}
    \caption{\textbf{Sighted rater prompt comparisons}. Mean Likert scores across five (of 11) key dimensions ($\uparrow$ indicates higher is better). Sighted raters evaluated videos played with AD generated from different prompts.}
    \label{fig:sighted_prompt_comparison}
\end{figure*}

This section presents the detailed inter-annotator correlation and scoring statistics for our Sighted Rater study (\cref{tab:inter_annotator_agreement_full}) alongside the sighted rater evaluation across prompting strategies (\cref{fig:sighted_prompt_comparison}).
\begin{table}[!ht]
\centering
\resizebox{0.5\columnwidth}{!}{%
\begin{tabular}{lrr}
\toprule
Metric & Krippendorff's $\alpha$ & Avg Std Dev \\
\midrule
Timeliness & 0.721 & 0.581 \\
Audio Issues & 0.700 & 0.545 \\
Objective & 0.665 & 0.437 \\
Gaps of Silence & 0.647 & 0.637 \\
Overall Quality & 0.581 & 0.681 \\
Descriptive & 0.579 & 0.699 \\
Relevance & 0.563 & 0.641 \\
Accuracy & 0.547 & 0.706 \\
Clarity & 0.544 & 0.709 \\
Coverage & 0.539 & 0.700 \\
Consistency/ Redundancy & 0.537 & 0.696 \\
\bottomrule
\end{tabular}
}
\caption{Inter-Annotator Agreement (Krippendorff's $\alpha$) and Average Standard Deviation per Metric. Overall (all metrics): 0.654 }
\label{tab:inter_annotator_agreement_full}
\end{table}

\section{Full Distribution of Scores and Qualitative Insights from the BLV User Study}
\begin{figure*}[!h]
    \centering
    \includegraphics[width=\linewidth]{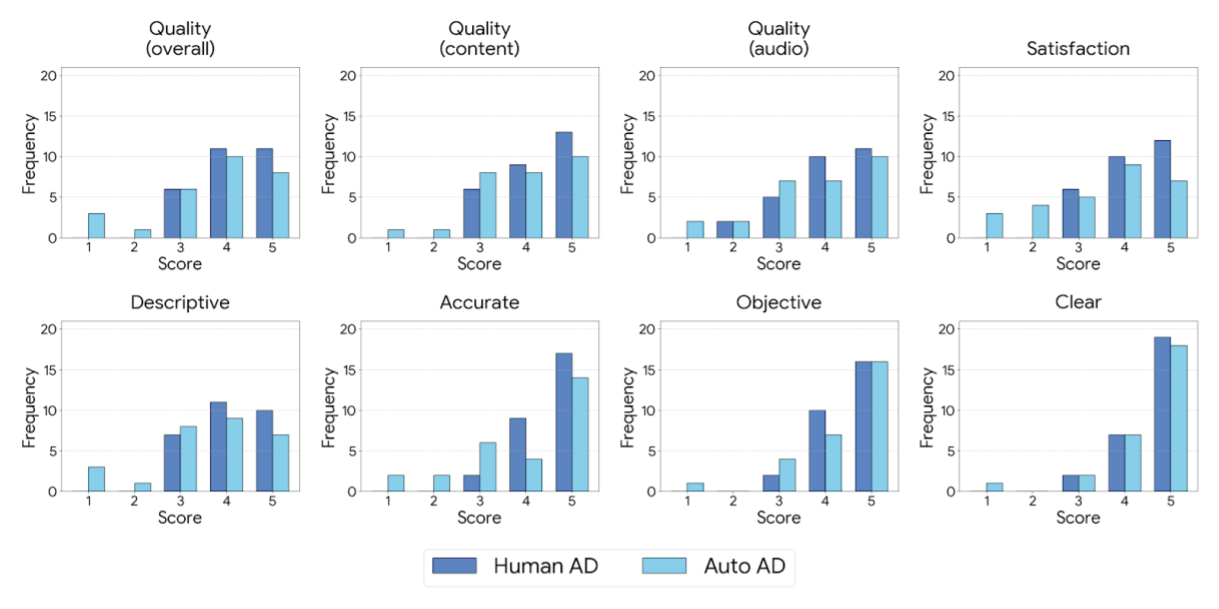}
    \caption{\textbf{BLV study.} Distribution of Likert scores across 8 key dimensions. BLV participants evaluated videos played with human written ADs and AutoAD (Gemini-2.5 Flash with Gd. + Timing ASR). They rated both AD versions consistently high (ratings of 4 or 5) across all metrics. A few of the videos with automated AD was rated lower with respect to audio quality, and in terms of overall quality and satisfaction when compared to the same videos played with human generated AD.}
    \label{fig:placeholder}
\end{figure*}

\begin{figure}[h]
    \centering
    \vspace{-0.cm}
    \includegraphics[width=0.6\linewidth]{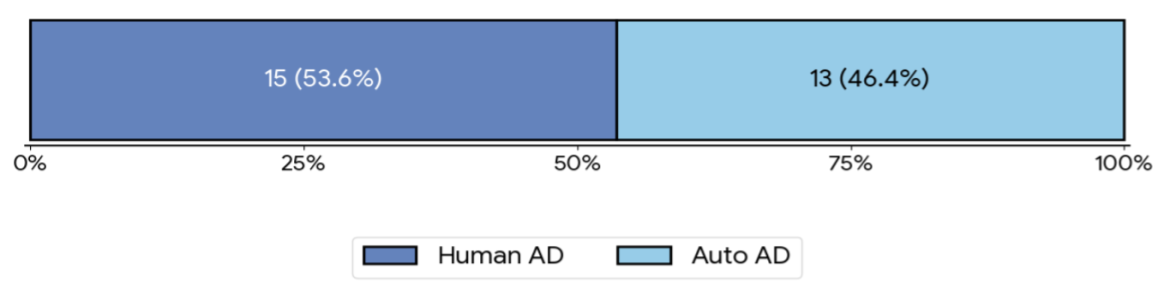}
    \caption{\textbf{BLV Study.} In a blind study, BLV participants listening to the same video with human generated AD and AutoAD (Gemini-2.5 Flash with Gd. + Timing ASR) when asked to pick which AD version they liked better, marginally preferred the video version with human generated AD.}
    \label{fig:blv_pref}
    \vspace{-0.2cm}
\end{figure}

This section visually documents the rating patterns observed across the blind and low-vision cohort evaluation (\cref{fig:placeholder,fig:blv_pref}) and summarizes qualitative participant feedback.

\subsection{BLV Qualitative Feedback and Individual Preference Variance}
\label{appx_sec:blv_qualitative}
Consistent with prior accessibility literature~\cite{natalie2024audio,liu2021what,jiang2024s}, post-task interviews with our 7 BLV participants ($28$ paired video evaluations) revealed substantial individual-level and genre-dependent variance in AD preferences. While professional human AD was favored overall ($53.6\%$) for its smooth narrative flow, concise phrasing, and natural integration around dialogue pauses, AutoAD (\textit{Gemini-2.5 Flash} with \textit{Gd. + Timing ASR}) was preferred in $46.4\%$ of comparisons---particularly on visually dense travel, nature, and instructional videos where participants valued exhaustive enumeration of visual objects, settings, and on-screen text. Conversely, when AutoAD was penalized, participants cited occasional redundancy with sound effects or abrupt sentence truncation when an AD segment reached the onset of video speech.

\newpage
\section{Extended Empirical Metric Validation and Reference-Free Evaluation}
\label{appx_sec:metric_validation}
This section provides the full empirical correlation tables between our automated evaluation metrics and human judgments across the Sighted Rater cohort ($n = 42$ videos, \cref{tab:sighted_metric_corr_full}), the BLV Participant cohort ($n = 28$ video evaluations, \cref{tab:blv_metric_corr_full}), and our Reference-Free \texttt{Quality-RF} comparison (\cref{tab:rf_vs_ref_corr_full}).

\begin{table*}[!ht]
\centering
\caption{\textbf{Automated Metrics vs.\ Sighted Rater Judgments ($n = 42$ videos).} Video-level Spearman rank correlations ($\rho$) and $p$-values alongside qualitative interpretations.}
\label{tab:sighted_metric_corr_full}
\resizebox{\textwidth}{!}{%
\begin{tabular}{@{}lllrrp{6.5cm}@{}}
\toprule
\textbf{Automated Metric} & \textbf{Metric Type} & \textbf{Target Human Dimension} & \textbf{Spearman $\rho$} & \textbf{$p$-value} & \textbf{Interpretation} \\
\midrule
\textbf{Audio Overlaps (G7)} & Programmatic (TTS + ASR) & Audio Issues (No Overlaps) & $\mathbf{-0.467}$ & $\mathbf{0.0018^{**}}$ & \textbf{Moderate alignment} (fewer overlap seconds $\rightarrow$ higher human audio rating) \\
\textbf{Avg.\ Silence Duration (G8)} & Programmatic (TTS + ASR) & Gaps of Silence (Few Quiet Gaps) & $\mathbf{-0.415}$ & $\mathbf{0.0063^{**}}$ & \textbf{Moderate alignment} (shorter silent gaps $\rightarrow$ higher human rating) \\
\textbf{VideoA11y-Ref Clear} & LLM Judge & Clarity & $\mathbf{+0.453}$ & $\mathbf{0.0026^{**}}$ & \textbf{Moderate positive alignment} \\
\textbf{VideoA11y-Ref Descriptive} & LLM Judge & Descriptive & $\mathbf{+0.328}$ & $\mathbf{0.0338^{*}}$ & \textbf{Statistically significant positive alignment} \\
\textbf{Prioritization (G1.1)} & LLM Judge (Visual Info) & Coverage & $+0.246$ & $0.1160$ & Positive trend \\
\textbf{Timeliness (G2)} & LLM Judge (Visual Info) & Timeliness & $-0.228$ & $0.1470$ & Expected negative trend (lower offset seconds $\rightarrow$ higher human rating) \\
\bottomrule
\end{tabular}%
}
\end{table*}

\begin{table*}[!ht]
\centering
\caption{\textbf{Automated Metrics vs.\ BLV Listener Judgments ($n = 28$ video evaluations).} Video-level Spearman rank correlations ($\rho$) and $p$-values.}
\label{tab:blv_metric_corr_full}
\resizebox{\textwidth}{!}{%
\begin{tabular}{@{}lllrrp{6.5cm}@{}}
\toprule
\textbf{Automated Metric} & \textbf{Metric Type} & \textbf{Target BLV Dimension} & \textbf{Spearman $\rho$} & \textbf{$p$-value} & \textbf{Interpretation / Note} \\
\midrule
\textbf{VideoA11y-Ref Clear} & LLM Judge (Ref-based) & Clarity & $\mathbf{+0.388}$ & $\mathbf{0.0411^{*}}$ & \textbf{Statistically significant positive alignment} \\
\textbf{No Redundancy (G1.2)} & LLM Judge (Ref-free) & Consistency / Non-Redundancy & $\mathbf{+0.386}$ & $\mathbf{0.0427^{*}}$ & \textbf{Statistically significant positive alignment} \\
\textbf{VideoA11y-Ref Descriptive} & LLM Judge (Ref-based) & Descriptive & $+0.340$ & $0.0771$ & Marginal positive alignment ($p < 0.10$) \\
\textbf{VideoA11y-Ref Objective} & LLM Judge (Ref-based) & Objective & $+0.335$ & $0.0811$ & Marginal positive alignment ($p < 0.10$) \\
\textbf{Audio Overlaps (G7)} & Programmatic (Ref-free) & Audio Quality & $+0.132$ & $0.5024$ & Low correlation because the study player truncated AD at speech onset \\
\textbf{VideoA11y-Ref Accurate} & LLM Judge (Ref-based) & Perceived Accuracy & $-0.264$ & $0.1743$ & Negative correlation due to reference-matching bias (resolved by \texttt{Quality-RF}) \\
\bottomrule
\end{tabular}%
}
\end{table*}

\begin{table*}[!ht]
\centering
\caption{\textbf{Reference-Free (\texttt{Quality-RF}) vs.\ Reference-Based (\texttt{VideoA11y-Ref}) Correlation with BLV Human Ratings ($n = 28$).} Replacing the human reference text with raw video and ASR transcript reverses the negative accuracy correlation and strengthens alignment across all four rubric dimensions.}
\label{tab:rf_vs_ref_corr_full}
\resizebox{\textwidth}{!}{%
\begin{tabular}{@{}lrrrrp{6.2cm}@{}}
\toprule
\textbf{VideoA11y Dimension} & \textbf{Ref-Based $\rho$} & \textbf{Ref-Based $p$} & \textbf{Ref-Free $\rho$ (Video + ASR)} & \textbf{Ref-Free $p$} & \textbf{Key Finding} \\
\midrule
\textbf{Descriptive} & $+0.340$ & $0.0771$ & $\mathbf{+0.583}$ & $\mathbf{0.0011^{**}}$ & \textbf{+0.243 $\rho$ gain; highly significant} \\
\textbf{Clear} & $\mathbf{+0.388}$ & $\mathbf{0.0411^{*}}$ & $\mathbf{+0.415}$ & $\mathbf{0.0281^{*}}$ & \textbf{+0.027 $\rho$ gain; statistically significant} \\
\textbf{Objective} & $+0.335$ & $0.0811$ & $+0.365$ & $0.0561$ & \textbf{+0.030 $\rho$ gain; strong marginal trend} \\
\textbf{Accurate} & $-0.264$ & $0.1743$ & $+0.244$ & $0.2109$ & \textbf{Reverses negative correlation (+0.508 swing)} \\
\bottomrule
\end{tabular}%
}
\end{table*}

\textbf{Ablation of Reference-Free Input Representations.} We evaluated four reference-free input configurations for \texttt{Quality-RF}: (1) \textit{video-only}, (2) \textit{video + ASR transcript}, (3) \textit{dense visual info + ASR}, and (4) \textit{video + dense visual info + ASR}. We found that providing the raw \textbf{video + ASR transcript} (\cref{tab:rf_vs_ref_corr_full}, \cref{fig:prompt_videoa11y_rf}) achieves the strongest overall correlation across all four dimensions ($\rho = +0.583$ on Descriptive, $\rho = +0.415$ on Clear, and $\rho = +0.244$ on Accurate), whereas text-only \textit{dense visual info + ASR} achieves the highest standalone Accuracy correlation ($\rho = +0.289$) when raw video frames cannot be passed to the judge.

\textbf{TTS Speaking Rate.} Our temporal metrics (\textbf{G7}, \textbf{G8}) render AD segments with an off-the-shelf TTS engine at a fixed $1.2\times$ speaking rate. Because a segment's spoken duration scales inversely with the rate, faster speech shortens AD intervals and reduces overlap seconds, and slower speech lengthens them; the absolute values of these metrics therefore depend on the chosen rate. We do not study other rates here. Faster or adaptive per-gap speaking rates are a promising direction, as they would allow more descriptions to fit into the available silences.

\newpage
\section{Questions for Sighted Raters}
\label{appx_sec:sighted_rater_qs}
This section documents the specific inquiry components provided to non-visually impaired participants. It lists each question and its corresponding categorical option to detail our human evaluation mechanics.
\begin{enumerate}[label=\textbf{\arabic*.}, leftmargin=*]

    \item \textbf{Coverage:} The descriptions covered the important visual information necessary for a blind user to understand the video.
    \begin{itemize}[label=$\square$]
        \item Strongly disagree. All important information was not covered.
        \item Slightly disagree. Most important information was not covered.
        \item Neither agree or disagree. Some important information was covered and some important information was missing.
        \item Slightly agree. Most important information was covered.
        \item Strongly agree. All important information was covered.
    \end{itemize}

    \item \textbf{Accuracy:} The descriptions were accurate.
    \begin{itemize}[label=$\square$]
        \item Strongly disagree. All of the information was inaccurate (potentially misleading).
        \item Slightly disagree. Most information was inaccurate.
        \item Neither agree or disagree. Some information was accurate and some was inaccurate.
        \item Slightly agree. Most information was accurate.
        \item Strongly agree. All information was accurate.
    \end{itemize}

    \item \textbf{Relevance:} The descriptions included relevant information.
    \begin{itemize}[label=$\square$]
        \item Strongly disagree. All of the information was irrelevant (not needed for a blind user to follow the video).
        \item Slightly disagree. Most information was irrelevant.
        \item Neither agree or disagree. Some information was relevant and some was irrelevant.
        \item Slightly agree. Most information was relevant.
        \item Strongly agree. All information was relevant.
    \end{itemize}

    \item \textbf{Consistency/Redundancy:} The descriptions account for information user already has from the video’s audio and prior descriptions.
    \begin{itemize}[label=$\square$]
        \item Strongly disagree. All descriptions repeated information the user would already know from the video’s audio or prior descriptions.
        \item Slightly disagree. Most descriptions repeated information the user would already know from the video’s audio or prior descriptions.
        \item Neither agree or disagree. Some of the descriptions provided new information but some of them included information the user would already know from the audio or prior descriptions.
        \item Slightly agree. Most descriptions provided new information but a few descriptions included information the user would already know from the video’s audio or prior descriptions.
        \item Strongly agree. All descriptions provided new information that supplemented the existing video’s audio and prior descriptions.
    \end{itemize}

    \item \textbf{Overall Quality:} Overall, the content of the descriptions was high quality and engaging to listen to without the visuals.
    \begin{itemize}[label=$\square$]
        \item Strongly disagree. All of the content of the narration had significant quality issues such that it was difficult to follow or not engaging to listen to. For example, the narration was overly repetitive and dull, or it had inconsistent references to characters, objects, or settings that made it difficult to follow.
        \item Slightly disagree. Most of the content of the narration had significant quality issues such that it was difficult to follow or not engaging to listen to.
        \item Neither agree or disagree. Some of the content of the narration was easy to follow and engaging to listen to. Some moments had quality issues that made the narration difficult to follow or not engaging to listen to.
        \item Slightly agree. Most of the content of the narration was easy to follow and engaging to listen to. The language of the narration was mostly clear, illustrative, and engaging.
        \item Strongly agree. All of the content of the narration was easy to follow and engaging to listen to. The language of the narration was entirely clear, illustrative, and engaging.
    \end{itemize}

    \item \textbf{Clarity:} The descriptions were clear and easy to follow considering the audio alone.
    \begin{itemize}[label=$\square$]
        \item Strongly disagree. All descriptions were unclear or hard to follow when considering the information in the audio alone (e.g., used unclear or inconsistent references to characters, settings, or objects).
        \item Slightly disagree. Most descriptions were unclear or hard to follow, but some descriptions were easy to follow when considering the information in the audio alone.
        \item Neither agree or disagree. Some of the descriptions were clear and easy to follow, and some were not clear or hard to follow when considering the information in the audio alone.
        \item Slightly agree. Most of the descriptions were clear and easy to follow when considering the information in the audio alone.
        \item Strongly agree. All descriptions were clear and easy to follow from the audio alone.
    \end{itemize}

    \item \textbf{Descriptive:} The descriptions were descriptive in that they provided vivid details about objects, people, and settings while maintaining a concise narrative flow.
    \begin{itemize}[label=$\square$]
        \item Strongly disagree. All of the descriptions were not descriptive. Instead, they were dull or repetitive, providing little useful information.
        \item Slightly disagree. Most of the descriptions were not descriptive, but instead they were dull or repetitive.
        \item Neither agree or disagree. Some of the descriptions included vivid details and some were dull or repetitive.
        \item Slightly agree. Most of the descriptions included vivid details and the description mostly maintained a concise narrative flow.
        \item Strongly agree. All of the descriptions included vivid details while maintaining a concise narrative flow.
    \end{itemize}

    \item \textbf{Objective:} The descriptions were objective in that they report what is visible without adding personal opinions or assumptions (e.g., about character’s relationships, or identity attributes).
    \begin{itemize}[label=$\square$]
        \item Strongly disagree. All of the descriptions contained subjective opinions or assumptions.
        \item Slightly disagree. Most of the descriptions included subjective opinions or assumptions.
        \item Neither agree or disagree. Some of the descriptions were objective but some of the descriptions included subjective opinions or assumptions.
        \item Slightly agree. Most of the descriptions were objective but some were subjective.
        \item Strongly agree. All of the descriptions were objective.
    \end{itemize}

    \item \textbf{Timeliness:} The descriptions appeared close to the time the event occurred in the video (usually within 3-5s).
    \begin{itemize}[label=$\square$]
        \item Strongly disagree. All events were not described close to the time they occurred.
        \item Slightly disagree. Most events were not described close to the time they occurred.
        \item Neither agree or disagree. Some events were described close to the time they occurred.
        \item Slightly agree. Most events were described close to the time they occurred.
        \item Strongly agree. All events were described close to the time they occurred.
    \end{itemize}

    \item \textbf{Audio Issues:} All descriptions were clear to listen to and did not have audio issues such as overlapping speech.
    \begin{itemize}[label=$\square$]
        \item Strongly disagree. All descriptions had audio issues such as overlapping speech. It is difficult to listen to the descriptions due to audio issues.
        \item Slightly disagree. Most descriptions had audio issues such as overlapping speech.
        \item Neither agree or disagree. There were some descriptions that were clear and some that would be difficult to listen to due to audio issues.
        \item Slightly agree. Most descriptions were clear without audio issues.
        \item Strongly agree. All descriptions were clear without audio issues.
    \end{itemize}

    \item \textbf{Gaps of Silence:} There were not large gaps of silence in the descriptions (6-10s quiet gaps).
    \begin{itemize}[label=$\square$]
        \item Strongly disagree. All gaps of silence in the video did not have descriptions.
        \item Slightly disagree. There were frequently long gaps of silence throughout the video.
        \item Neither agree or disagree. There were some long gaps of silence throughout the video.
        \item Slightly agree. There were long gaps of silence once or twice in the video.
        \item Strongly agree. There were no long gaps of silence.
    \end{itemize}

\end{enumerate}

\section{Examples showing Inline Audio Description Constraints}

\begin{figure*}[!ht]
    \centering
    \includegraphics[width=\linewidth]{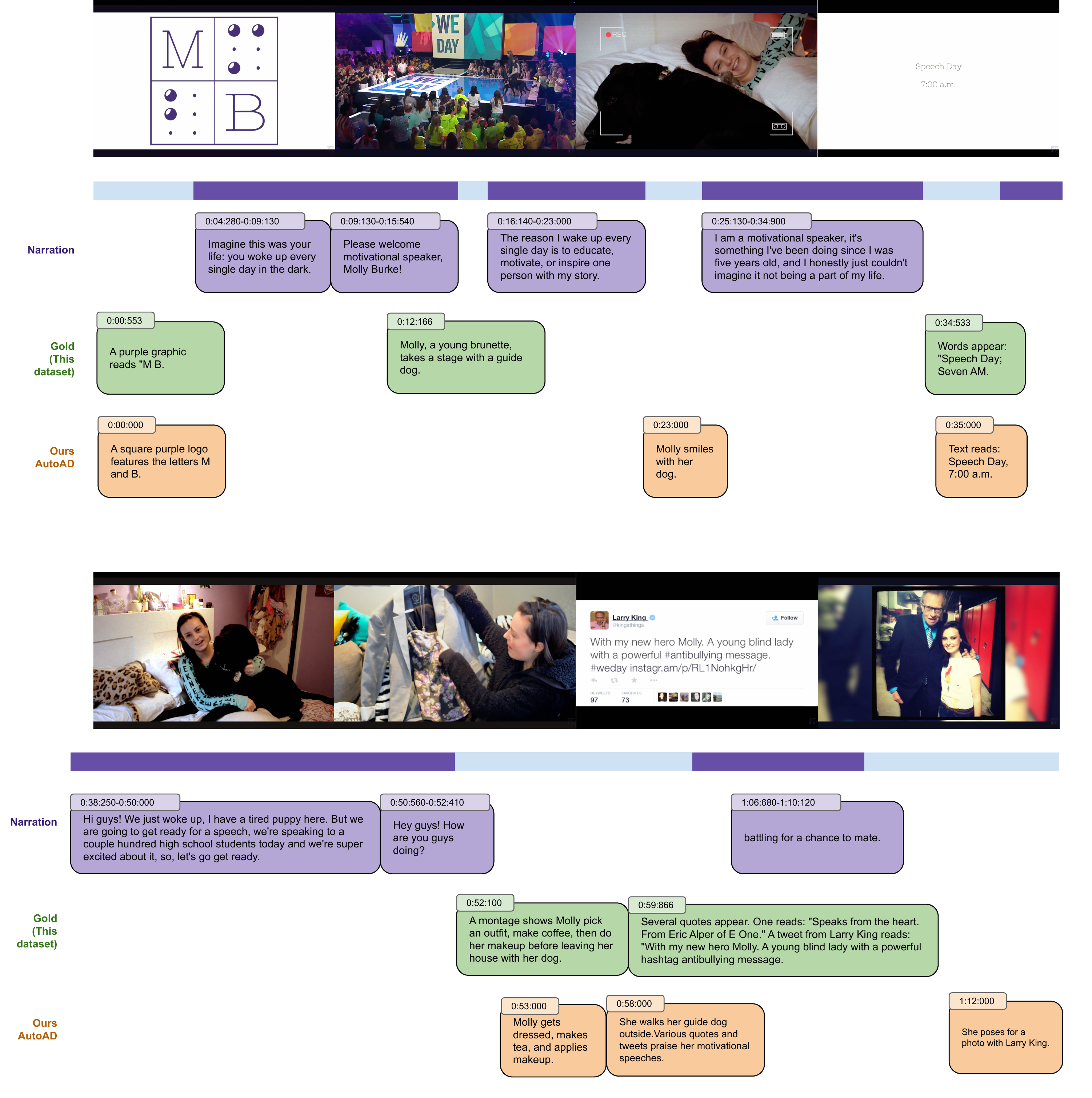}
    \caption{Example demonstrating more detailed descriptions from human annotators relative to model generated description despite the tight timing constraints of the video.}
    \label{fig:molly_1min}
\end{figure*}

\begin{figure*}[!ht]
    \centering
    \includegraphics[width=\linewidth]{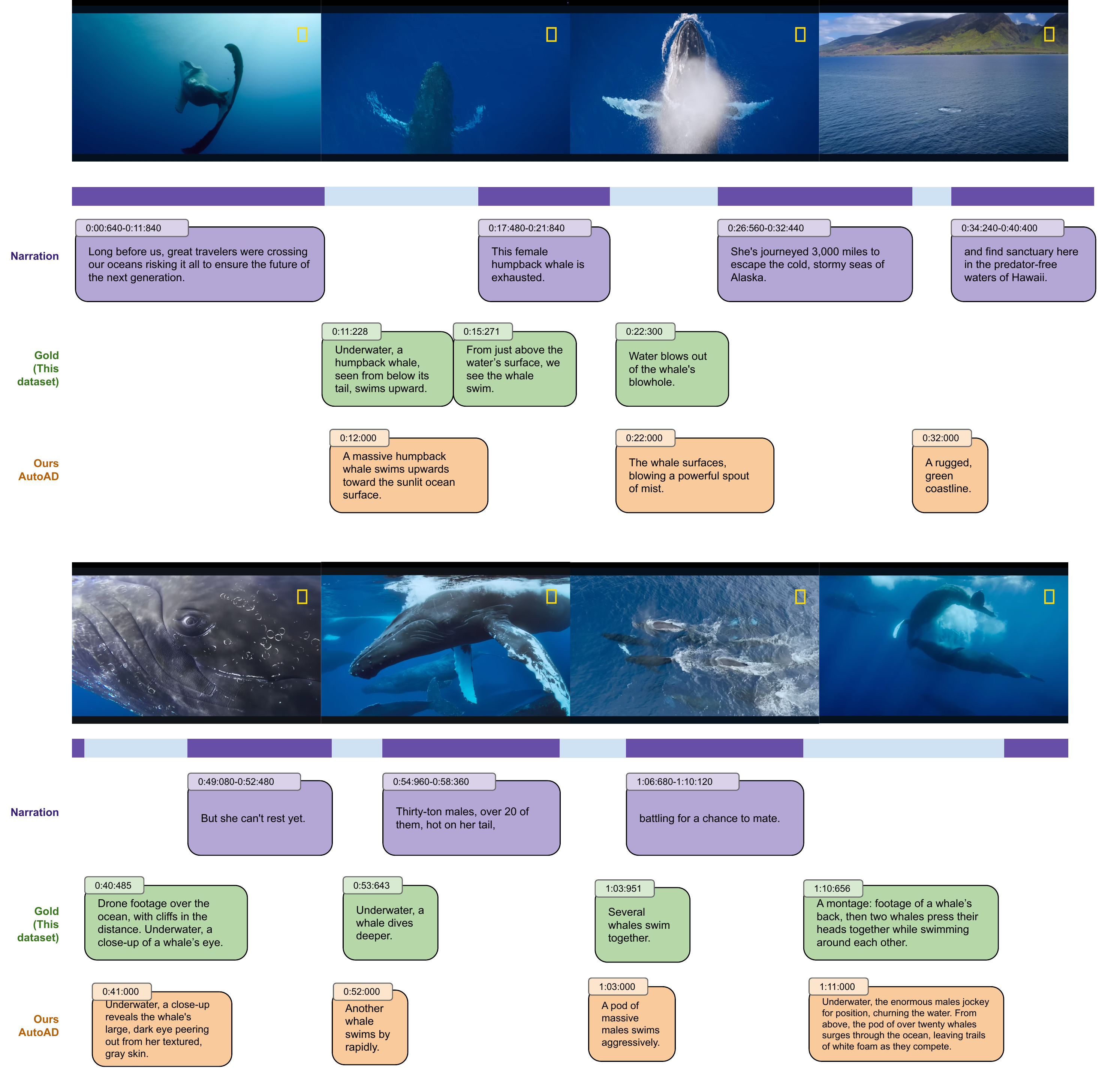}
    \caption{Example demonstrating human annotators describing the perspective of the video in addition to vivid descriptions going with the flow of the narration.}
    \label{fig:whale_1min}
\end{figure*}

\newpage

\section{Examples with different prompts}
This section showcases descriptive side-by-side comparisons taken from standard frames in our repository. It features illustrative contrasts between automatically produced scripts and those developed by our expert pairs.
\begin{figure*}[!tbh]
    \vspace{-0.4cm}
    \centering
    \includegraphics[width=\linewidth]{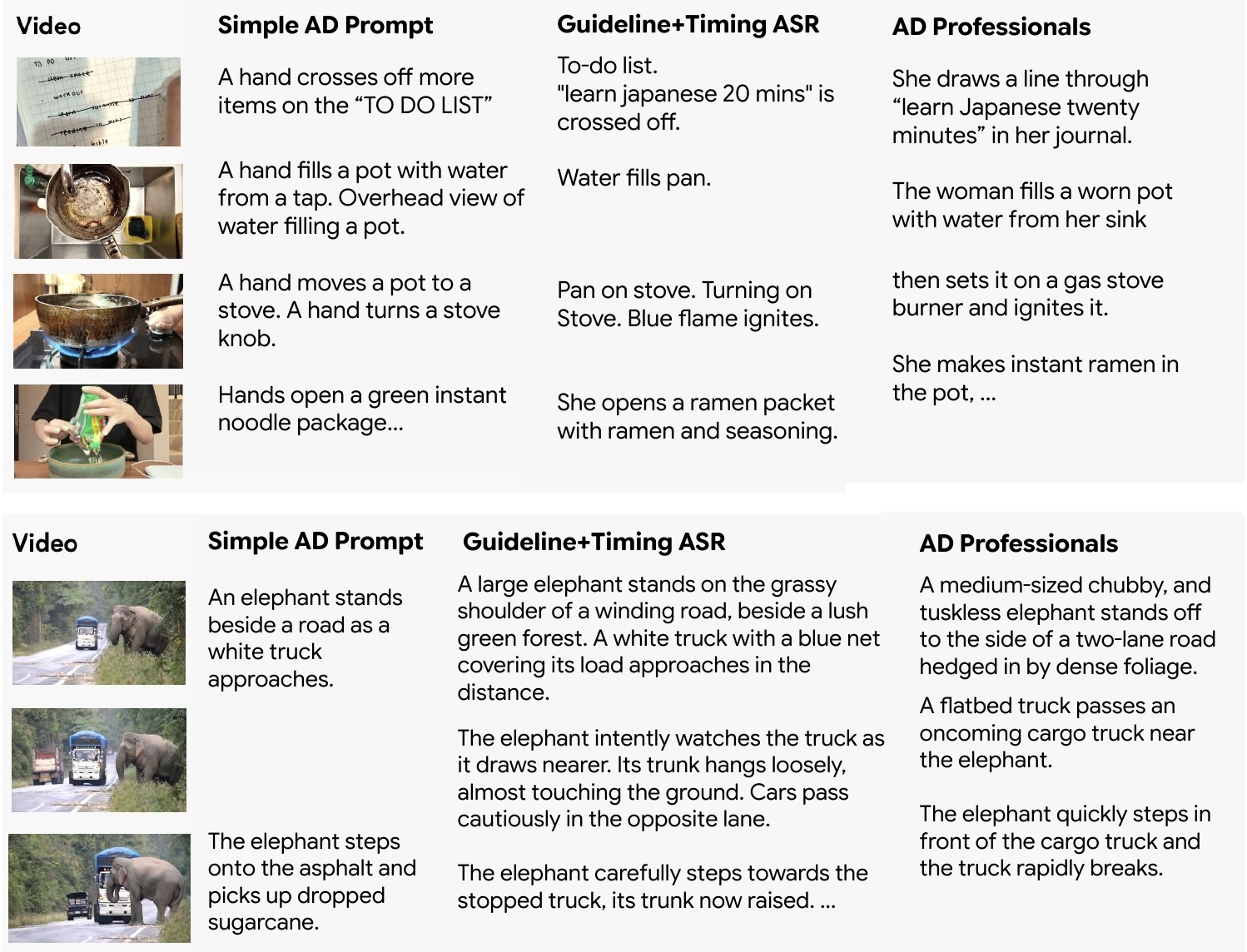}
    \caption{\textbf{Dataset Examples.} Model Generated AD (created with the Simple AD and Gd.\ + Timing ASR prompts) and corresponding professionally annotated AD for a video segment. Here, we illustrate each video segment with manually selected video frames. Top example demonstrates narrative flow in the professionally annotated AD (describes woman's story rather than just the actions), and the example below shows a missed event in the Model Generated ADs (the truck rapidly breaks for the elephant).}
    \label{fig:dataset_examples_appendix}
    \vspace{-0.5cm}
\end{figure*}

\newpage
\section{Prompts for AD Generation}
\label{appx_sec:prompts_ad_gen}
This section displays the actual systemic directives fed into multimodal architectures to elicit narrations. It archives variations of baseline instructions, transcript integrations, and rigorous timing constraints.

\begin{figure*}[!htb]
\begin{tcolorbox}[width=\linewidth,title=Visual Information Prompt]
\ttfamily\fontsize{9pt}{9.5pt}\selectfont %

Please watch this video and list out all the important visual events
that happen in the video narrative with timestamps. Please be very
descriptive with time stamps.

\paragraph{}The output format must be exactly like this:

\begin{description}
    \item * One scene description per line.
    \item * Each description follows the format <timestamp> <description>
    \begin{itemize}
        \item * The timestamp must be in the format 00:00:00.000 (hh:mm:ss.ms) followed by a "-".
    \end{itemize}
     \item * The descriptions must be sorted by timestamp in ascending order.
     \item * **CRUCIAL**: The descriptions must follow this format and cover the entire video.
\end{description}  

Example output:

00:00:00.000 visual\_event\_description1

00:00:07.123 visual\_event\_description2

00:00:23.745 visual\_event\_description3

\paragraph{}

Afterwards please watch the video and list all the text on screen and when it
appeared on the screen.

Please list all people in the video if there are any and what we know about
the people.

Finally, please provide a summary of the video narrative.

\end{tcolorbox}
\caption{Prompt used to generate detailed visual information in the video. The responses are used when evaluating the VLM and human generated descriptions.}
\label{fig:prompt_vis_info}
\end{figure*}

\begin{figure*}[!htb]
\begin{tcolorbox}[width=\linewidth,title=Simple Prompt]
\ttfamily\fontsize{9pt}{9.5pt}\selectfont %

You are a professional audio description expert. Create inline audio descriptions for the provided video to describe visual events only when there is no speech or dialog. For each audio description, provide the start time as a timestamp (e.g., 01:12:45.403 with HH:MM:SS.MS) then provide the text of the description.

\end{tcolorbox}
\caption{Most basic baseline prompt used for generating audio description.}
\label{fig:prompt_basic}
\end{figure*}

\newpage
\begin{center}
\begin{tcolorbox}[
    breakable,
    enhanced, %
    width=\linewidth,
    title=Guideline AD Prompt,
    fontupper=\ttfamily\fontsize{9pt}{9.5pt}\selectfont
]
You are an expert audio description writer.

Your task is to write high-quality audio descriptions for videos,
ensuring accessibility for visually impaired viewers.
Audio descriptions should narrate key visual elements
that are not already conveyed through the dialogue or other audio elements.

Please adhere to the following guidelines:

\paragraph{}
What to Describe:

\begin{description}[itemsep=0pt, parsep=0pt, topsep=0pt]
\item * Describe what you see. Focus on the essential visual information needed to understand the scene.
\item * Be concise.
\item * Always read on-screen text exactly as it appears.
\item * Be factual, objective, and avoid speculation.
\item * Use proper terminology and names whenever possible.
\item * All of the descriptions will need to fit in the space between dialog
  (assume 1-3 words per second when spoken aloud).
  If there is no space, then you need to skip the description.
\item * Try to match the mood of the video with your descriptions.
\item * Describe color only when it is vital to the comprehension of content.
\item ** CRUCIAL: ** Your descriptions must cover the entire video.
\end{description}

\paragraph{}
What NOT to Describe:

\begin{description}[itemsep=0pt, parsep=0pt, topsep=0pt]
\item * Do not talk over the dialog or any other essential audio.
\item * Do not describe what can be inferred from the audio.
\item * Do not over-describe; less is more. Keep descriptions brief and to the point.
\item * Do not interpret or editorialize. Stick to describing what is visually present.
\item * Do not give away secrets, surprises, or sight gags before they happen.
\item * Do not censor (sex, violence, gore, emotions), describe them accurately.
\item * Do not describe obvious sound cues such as dialogue, a phone ringing, or a dog barking.
\end{description}

\paragraph{}
Remember to only place descriptions between gaps of speech.
Strive for clear, concise, and informative audio descriptions
that enhance the viewing experience without distracting from the original content.

  The output format must be a valid JSON with a single field called "response" containing the following:
  \begin{description}
      \item * One scene description per line.

      \item * Each description follows the format <timestamp>-STANDARD <description>
      \begin{itemize}
        \item * The timestamp must be in seconds in the format 0:00:00.000 (h:mm:ss.ms) followed by a "-"
        \item * The type of description should always be STANDARD
        \item * The description must be in a single line.
        \end{itemize}
      \item * The descriptions must be sorted by timestamp in ascending order.
      \item * **CRUCIAL**: The descriptions must follow this format and cover the entire video.
      \item * **CRUCIAL**: The format must be valid.
  \end{description}

\end{tcolorbox}
\captionof{figure}{Prompt used to generate audio description with just the guidelines and no video transcripts.}
\label{fig:prompt_guideline_ad}
\end{center}
\begin{center}
\begin{tcolorbox}[
    breakable,
    enhanced, %
    width=\linewidth,
    title=Guideline AD and ASR Prompt,
    fontupper=\ttfamily\fontsize{9pt}{9.5pt}\selectfont
]
You are an expert audio description writer.

Your task is to write high-quality audio descriptions for videos,
ensuring accessibility for visually impaired viewers.
Audio descriptions should narrate key visual elements
that are not already conveyed through the dialogue or other audio elements.

The video transcript with timestamps (start and end time in seconds) is
provided below. Make sure to generate descriptions only in between dialogs
where there is no speech in the video. The descriptions should fit in the 
gaps when spoken out loud. (assume you can fit in 1-3 words per second)

\paragraph{}

<asr\_transcript>

ASR\_TRANSCRIPT\_GOES\_HERE\_IF\_AVAILABLE

</asr\_transcript>

\paragraph{}
Please adhere to the following guidelines:

\paragraph{}
What to Describe:

\begin{description}[itemsep=0pt, parsep=0pt, topsep=0pt]
\item * Describe what you see. Focus on the essential visual information needed to understand the scene.
\item * Be concise.
\item * Always read on-screen text exactly as it appears.
\item * Be factual, objective, and avoid speculation.
\item * Use proper terminology and names whenever possible.
\item * All of the descriptions will need to fit in the space between dialog
  (assume 1-3 words per second when spoken aloud).
  If there is no space, then you need to skip the description.
\item * Try to match the mood of the video with your descriptions.
\item * Describe color only when it is vital to the comprehension of content.
\item ** CRUCIAL: ** Your descriptions must cover the entire video.
\end{description}

What NOT to Describe:

\begin{description}[itemsep=0pt, parsep=0pt, topsep=0pt]
\item * Do not talk over the dialog or any other essential audio.
\item * Do not describe what can be inferred from the audio.
\item * Do not over-describe; less is more. Keep descriptions brief and to the point.
\item * Do not interpret or editorialize. Stick to describing what is visually present.
\item * Do not give away secrets, surprises, or sight gags before they happen.
\item * Do not censor (sex, violence, gore, emotions), describe them accurately.
\item * Do not describe obvious sound cues such as dialogue, a phone ringing, or a dog barking.
\end{description}

\paragraph{}
Remember to only place descriptions between gaps of speech.
Strive for clear, concise, and informative audio descriptions
that enhance the viewing experience without distracting from the original content.

  The output format must be a valid JSON with a single field called "response" containing the following:
  \begin{description}
      \item * One scene description per line.

      \item * Each description follows the format <timestamp>-STANDARD <description>
      \begin{itemize}
        \item * The timestamp must be in seconds in the format 0:00:00.000 (h:mm:ss.ms) followed by a "-"
        \item * The type of description should always be STANDARD
        \item * The description must be in a single line.
        \end{itemize}
      \item * The descriptions must be sorted by timestamp in ascending order.
      \item * **CRUCIAL**: The descriptions must follow this format and cover the entire video.
      \item * **CRUCIAL**: The format must be valid.
  \end{description}
\end{tcolorbox}
\captionof{figure}{Prompt used to generate audio description with the audio transcripts and the guidelines.}
\label{fig:prompt_asr_guideline}
\end{center}

\newpage
\begin{center}
\begin{tcolorbox}[
    breakable,
    enhanced, %
    width=\linewidth,
    title=Guideline + Timing ASR Prompt,
    fontupper=\ttfamily\fontsize{9pt}{9.5pt}\selectfont
]
You are an expert audio description writer.

Your task is to write high-quality audio descriptions for videos,
ensuring accessibility for visually impaired viewers.
Audio descriptions should narrate key visual elements
that are not already conveyed through the dialogue or other audio elements.

The video transcript with timestamps (start and end time in seconds) is
provided below. You MUST only generate descriptions in the silent gaps BETWEEN dialogue segments indicated by the timestamps.

\paragraph{}

<asr\_transcript>

ASR\_TRANSCRIPT\_GOES\_HERE\_IF\_AVAILABLE

</asr\_transcript>

\paragraph{}
**CRUCIAL TIMING \& LENGTH RULES TO THINK AND FOLLOW:**
\begin{description}[itemsep=0pt, parsep=0pt, topsep=0pt]
\item 1.  **No Overlap:** Descriptions MUST NOT overlap with any spoken dialogue in the provided transcript. Use the timestamps to find silent gaps.
\item 2.  **Calculate Gap Duration:** For each gap, calculate the available time in seconds.
\item 3.  **Strict Word Limit:** The number of words in any description MUST NOT exceed the gap duration (in seconds) multiplied by 3. This means a maximum of 3 words per second of available gap time.
\item 4.  **Sentence Limit:** Each description associated with a single timestamp MUST NOT exceed **three** sentences.
\item 5.  **Fit the Gap:** The spoken description must comfortably fit within the calculated silent gap, respecting both word and sentence limits.
\item 6.  **Skip if No Space:** If a visual element cannot be described within the available time and sentence limits in any gap, you MUST skip describing it.
\item 7.  **Silent videos:** If the video is a silent video with no speech (just music) then add more descriptions to cover the entire video.
\end{description}

\paragraph{}
Please adhere to the following guidelines:

\paragraph{}
What to Describe:

\begin{description}[itemsep=0pt, parsep=0pt, topsep=0pt]
\item * Describe what you see. Focus on the essential visual information needed to understand the scene.
\item * Be concise and fit the description within the time and sentence constraints.
\item * Always read on-screen text exactly as it appears.
\item * Be factual, objective, and avoid speculation.
\item * Use proper terminology and names whenever possible.
\item * Try to match the mood of the video with your descriptions.
\item * Describe color only when it is vital to the comprehension of content.
\item ** CRUCIAL: ** Your descriptions must cover the entire video, respecting the timing gaps.
\end{description}

What NOT to Describe:

\begin{description}[itemsep=0pt, parsep=0pt, topsep=0pt]
\item * Do not talk over the dialog or any other essential audio.
\item * Do not describe what can be inferred from the audio.
\item * Do not over-describe; less is more. Keep descriptions brief and to the point.
\item * Do not interpret or editorialize. Stick to describing what is visually present.
\item * Do not give away secrets, surprises, or sight gags before they happen.
\item * Do not censor (sex, violence, gore, emotions), describe them accurately.
\item * Do not describe obvious sound cues such as dialogue, a phone ringing, or a dog barking.
\end{description}

\paragraph{}
Remember to only place descriptions between gaps of speech.
Strive for clear, concise, and informative audio descriptions
that enhance the viewing experience without distracting from the original content.

  The output format must be a valid JSON with a single field called "response" containing the following:
  \begin{description}
      \item * One scene description per line.

      \item * Each description follows the format <timestamp>-STANDARD <description>
      \begin{itemize}
        \item * The timestamp must be in seconds in the format 0:00:00.000 (h:mm:ss.ms) indicating the START time of the description within a silent gap, followed by a "-"
        \item * The type of description should always be STANDARD
        \item * The description must be in a single line, containing no more than three sentences.
        \end{itemize}
      \item * The descriptions must be sorted by timestamp in ascending order.
      \item * **CRUCIAL**: The descriptions must follow this format and cover the entire video.
      \item * **CRUCIAL**: The format must be valid.
  \end{description}
\end{tcolorbox}
\captionof{figure}{Prompt used to generate audio description with the guidelines, the timestamped transcript, and the duration of each available gap and the time needed to speak (Gd.\ + Timing ASR).}
\label{fig:prompt_timing_asr}
\end{center}

\section{Prompts for AD Evaluation Metrics}
\label{appx_sec:prompts_ad_eval}
This section exhibits the specific prompts leveraged to operate LLM-as-a-judge scoring functions. It logs queries responsible for testing parameters like redundancy checks, content coverage, and narration accuracy.

\begin{figure*}[!htb]
\begin{tcolorbox}[width=\linewidth,title=Prioritization Prompt]
\ttfamily\fontsize{9pt}{9.5pt}\selectfont %

  You are evaluating if audio descriptions (AD) are covering visual information
  necessary to understand the video.
  For each line in the first AD output,
  score if it is covered in the visual information (0 - not covered, 1 - fully covered).

  Then, output only an array of all the scores like [0,0.5,1.0,1.0,0] 

\paragraph{}
First AD Output:

\texttt{<ad\_output>}

\paragraph{}
  Second  visual information of the video:

\texttt{<dense\_visual\_info>}

\end{tcolorbox}
\caption{Prompt used to evaluate the prioritization of important events in the audio description based on detailed visual information.}
\label{fig:prompt_coverage}
\end{figure*}

\begin{figure*}[!htb]
\begin{tcolorbox}[width=\linewidth,title=On-Screen Text Narration Prompt]
\ttfamily\fontsize{9pt}{9.5pt}\selectfont %
I will provide the on-screen text that occurs in the video. For each on-screen text segment, check if the text is reported in either the transcript or audio description at a nearby time. If a text chunk is not said in the transcript or audio description, count the number of unsaid words in the chunk. Then, return a list of unsaid word counts e.g., [0, 0, 1, 0, 5, 0]. ONLY RETURN THE LIST.

\paragraph{}
Here is the interleaved transcript labeled [TRANSCRIPT] and audio descriptions labeled [AD]:

\texttt{<interleaved\_transcript\_ad\_str>}

\paragraph{}
Here is everything visual that occurred in the video including ON-SCREEN TEXT:

\texttt{<visual\_info\_str>}
\end{tcolorbox}
\caption{\textit{Initial version} of the On-Screen Text (G1.3) metric, which used an LLM judge to count unsaid on-screen words. All reported G1.3 results instead use a programmatic word-level recall (\cref{sec:content_metrics}): the fraction of on-screen-text words listed in the dense visual information that appear anywhere in the AD or the ASR transcript; no LLM judge is involved.}
\label{fig:prompt_onscreen}
\end{figure*}

\begin{figure*}[!htb]
\begin{tcolorbox}[width=\linewidth,title=Redundancy Prompt]
\ttfamily\fontsize{9pt}{9.5pt}\selectfont %
I will provide the video transcript [TRANSCRIPT] and audio description [AD] interleaved, such that they are in order as they would be played in the video.

Then, for each audio description [AD], check if the audio description repeats the nearby transcript [TRANSCRIPT] or would otherwise be obvious from the audio alone (e.g., describes offscreen door slam). Score each AD between 0 and 1 where no repeat - 0, slight repeat - 0.5, and egregious repeat - 1.0. The purpose is to remove unnecessary redundancy with the audio track from the audio description. Then, return a list of these assigned numbers e.g., [0, 0, 1.0, 0, 0, 0.5, 0.6]. ONLY RETURN THE LIST.

\paragraph{}
Here is the video transcript and audio description interleaved:

\texttt{<interleaved\_transcript\_ad\_str>}
\end{tcolorbox}
\caption{Prompt used to evaluate the redundancy of audio descriptions with the original transcript.}
\label{fig:prompt_redundancy}
\end{figure*}

\begin{figure*}[!htb]
\begin{tcolorbox}[width=\linewidth,title=Timeliness Prompt (Visual Information-based)]
\ttfamily\fontsize{9pt}{9.5pt}\selectfont %
    For each audio description item, identify the corresponding
    visual information if one exists, and report the time difference (in secs):

     \paragraph{}
 
    \noindent [audio description time] audio description
    
    [visual information time] visual information
    
    **[audio description time - visual info time]**
  
    The response should not contain any thing else.
  
    For example:
    -----
    \paragraph{}
    [00:03:01.183] Ants crawl on banana pieces.

    [00:03:12.257] Ants crawl on pieces of banana on rocks.
    
    **[11]**
    \paragraph{}
  
    [199] A light brown cockroach hangs from a screen, shedding its exoskeleton.
    
    No audio description.
    \paragraph{}
    
    **[N/A]**
    
    \paragraph{}
    -----
    \paragraph{}
  
    Here are the audio descriptions and visual information:
    \paragraph{}
  
    Audio descriptions (timestamp isin hh:mm:ss.ms when the audio description starts in the video):

    \paragraph{}
    
    \texttt{<audio\_descriptions>}
  
    Visual information (timestamp isin hh:mm:ss.ms indicate when the visual appeared in the video):
    
    \paragraph{}

    \texttt{<visual\_info>}

\end{tcolorbox}
\caption{Prompt for evaluating the timeliness of audio descriptions by comparing them to pre-extracted visual information.}
\label{fig:prompt_timeliness}
\end{figure*}

\newpage
\begin{center}
\begin{tcolorbox}[
    breakable,
    enhanced, %
    width=\linewidth,
    title=VideoA11y Prompt,
    fontupper=\ttfamily\fontsize{9pt}{9.5pt}\selectfont
]
You are an expert evaluator with a deep understanding of video description quality, particularly for making content accessible to blind and low vision (BLV) individuals.

Your role is to assess and rate video descriptions based on specific metrics.

\paragraph{}
Task: I have two video descriptions: one is the ground truth, and the other is generated by a model. Additionally, I have four evaluation metrics: Descriptive, Objective, Accurate, and Clear.

\paragraph{}
Please evaluate the model-generated description using the above metrics. Provide a rating from 1 to 5 for each metric, and briefly explain the reasons for each rating. Metrics Definition:

\paragraph{}
1. Descriptive: A descriptive description should provide vivid details about objects, people, and settings while maintaining a concise narrative flow. It should include essential information about the appearance of individuals, such as their clothing, facial expressions, and actions, and visual properties of objects, such as color and shape. For example, "A smiling woman, wearing a loose white dress, types on a laptop in a softly lit room."

\paragraph{}
2. Objective: An objective description should report what is visible without adding personal opinions or assumptions. For instance, describe two people as "a woman and a man" without adding any relationship context unless it is mentioned. It should also avoid guessing personal attributes like racial or gender identities unless explicitly clear.

\paragraph{}
3. Accurate: An accurate description should aim for precision in describing visible elements without imagination. It should ensure that all details, such as colors and spatial arrangements, are reported correctly. For instance, "Blue sky with white clouds" instead of "White sky with blue clouds" if that is what appears. Additionally, it should avoid adding unnecessary details that do not contribute to a deeper understanding of the scene.

\paragraph{}
4. Clear: A clear description should present information in the videos in a way that is easy to follow for blind and low vision individuals. It should describe the object or character's properties before the actions or relationships with other objects or characters. For example, "A man wearing sunglasses plays the piano." Pronouns should only be used when it is clear who they refer to, and the description should include any on-screen text if it is central to understanding. For instance, "A man in a black polo shirt is speaking. He is in a studio setting with a red couch and a blue background featuring the text 'Talk of the Town'".

\paragraph{}
Input: -

\paragraph{}
Ground truth video description:

{\ttfamily
<ground\_truth\_desc>

GROUND\_TRUTH\_DESC\_GOES\_HERE\_IF\_AVAILABLE

</ground\_truth\_desc>
}

\paragraph{}
Model-generated video description:
{\ttfamily
<generated\_desc>

GENERATED\_DESC\_GOES\_HERE\_IF\_AVAILABLE

</generated\_desc>
}

\paragraph{}
Output Format: Return the result as a string format dictionary with 
the following structure:

\paragraph{}
{"Descriptive": [Rating out of 5], "Objective": [Rating out of 5], "Accurate": [Rating out of 5], "Clear": [Rating out of 5], "Reason": "Your brief explanation here"}
\end{tcolorbox}
\captionof{figure}{Prompt used to evaluate the Reference-Based VideoA11y (\texttt{VideoA11y-Ref}) metrics.}
\label{fig:prompt_videoa11y}

\end{center}

\begin{center}
\begin{tcolorbox}[
    breakable,
    enhanced,
    width=\linewidth,
    title=Reference-Free Quality-RF Prompt (Video + ASR),
    fontupper=\ttfamily\fontsize{9pt}{9.5pt}\selectfont
]
You are an expert evaluator with a deep understanding of audio description quality, particularly for making video content accessible to blind and low vision (BLV) individuals.

\paragraph{}
Your role is to assess and rate audio descriptions based on specific metrics. Audio descriptions should describe important visual events that are not already captured in the existing dialog or audio of the video. The provided ASR transcript represents the spoken content in the video.

\paragraph{}
Task: I have a video, an ASR transcript of the video, and a model-generated audio description. Additionally, I have four evaluation metrics: Descriptive, Objective, Accurate, and Clear.

\paragraph{}
Please evaluate the model-generated audio description using the above metrics by comparing it to the video and the ASR transcript.

\paragraph{}
Before providing the ratings, please provide a brief summary of the video and your rationale for the ratings.

\paragraph{}
Metrics Definition:

\paragraph{}
1. Descriptive: A descriptive description should provide vivid details about objects, people, and settings while maintaining a concise narrative flow. It should include essential information about the appearance of individuals, such as their clothing, facial expressions, and actions, and visual properties of objects, such as color and shape. For example, "A smiling woman, wearing a loose white dress, types on a laptop in a softly lit room."

\paragraph{}
2. Objective: An objective description should report what is visible without adding personal opinions or assumptions. For instance, describe two people as "a woman and a man" without adding any relationship context unless it is mentioned. It should also avoid guessing personal attributes like racial or gender identities unless explicitly clear.

\paragraph{}
3. Accurate: An accurate description should aim for precision in describing visible elements without imagination. It should ensure that all details, such as colors and spatial arrangements, are reported correctly. For instance, "Blue sky with white clouds" instead of "White sky with blue clouds" if that is what appears. Additionally, it should avoid adding unnecessary details that do not contribute to a deeper understanding of the scene.

\paragraph{}
4. Clear: A clear description should present information in the videos in a way that is easy to follow for blind and low vision individuals. It should describe the object or character's properties before the actions or relationships with other objects or characters. For example, "A man wearing sunglasses plays the piano." Pronouns should only be used when it is clear who they refer to, and the description should include any on-screen text if it is central to understanding. For instance, "A man in a black polo shirt is speaking. He is in a studio setting with a red couch and a blue background featuring the text 'Talk of the Town'".

\paragraph{}
Input:

\paragraph{}
ASR transcript:

{\ttfamily
<asr\_transcript>

ASR\_TRANSCRIPT\_GOES\_HERE\_IF\_AVAILABLE

</asr\_transcript>
}

\paragraph{}
Model-generated audio description:
{\ttfamily
<generated\_desc>

GENERATED\_DESC\_GOES\_HERE\_IF\_AVAILABLE

</generated\_desc>
}

\paragraph{}
Output Format: Return the result as a string format dictionary with the following structure. Please output the fields in this exact order:

\paragraph{}
\{"Summary": "Your brief summary of the video here", "Rationale": "Your rationale for the ratings here", "Descriptive": [Rating out of 5], "Objective": [Rating out of 5], "Accurate": [Rating out of 5], "Clear": [Rating out of 5]\}
\end{tcolorbox}
\captionof{figure}{Prompt used to evaluate the reference-free \texttt{Quality-RF} metrics (VideoA11y rubric) directly against the raw video and ASR transcript without requiring human ground-truth AD.}
\label{fig:prompt_videoa11y_rf}
\end{center}

\end{document}